\documentclass[10pt,journal,compsoc]{IEEEtran}

\usepackage{rotating}

\usepackage{amsmath}
\usepackage{amssymb}
\usepackage{subcaption}
\usepackage{algorithm}
\usepackage{algorithmic}
\usepackage{enumitem}
\usepackage{multirow}
\usepackage{epsfig}
\usepackage{graphicx}
\usepackage{color}
\usepackage[pagebackref=true,breaklinks=true,letterpaper=true,colorlinks,bookmarks=false]{hyperref}

\ifCLASSOPTIONcompsoc
  \usepackage[nocompress]{cite}
\else
  \usepackage{cite}
\fi

\begin{document}
\title{Warping the Residuals for Image Editing with StyleGAN}

\author{
Ahmet Burak Yildirim, Hamza Pehlivan, Aysegul Dundar

% \IEEEcompsocitemizethanks{
% % \IEEEcompsocthanksitem A. B. Yildirim and A.Dundar are with Bilkent University Computer Science, Turkey.
% % \IEEEcompsocthanksitem H.Pehlivan is with MPI, Germany.
% % \IEEEcompsocthanksitem  E-mail: adundar@cs.bilkent.edu.tr %
% % \IEEEcompsocthanksitem  E-mail: 
% %  a.yildirim@bilkent.edu.tr
% % \IEEEcompsocthanksitem Manuscript received xxx
% }
}

% The paper headers
% \markboth{IEEE TRANSACTIONS ON PATTERN ANALYSIS AND MACHINE INTELLIGENCE}%
% {Shell \MakeLowercase{\textit{et al.}}: Bare Demo of IEEEtran.cls for Computer Society Journals}

\IEEEtitleabstractindextext{
\begin{abstract}
StyleGAN models show editing capabilities via their semantically interpretable latent organizations which require successful GAN inversion methods to edit real images. Many works have been proposed for inverting images into StyleGAN's latent space. However, their results either suffer from low fidelity to the input image or poor editing qualities, especially for edits that require large transformations. That is because low-rate latent spaces lose many image details due to the information bottleneck even though it provides an editable space. On the other hand, higher-rate latent spaces can pass all the image details to StyleGAN for perfect reconstruction of images but suffer from low editing qualities. In this work, we present a novel image inversion architecture that extracts high-rate latent features and includes a flow estimation module to warp these features to adapt them to edits. The flows are estimated from StyleGAN features of edited and unedited latent codes. By estimating the high-rate features and warping them for edits, we achieve both high-fidelity to the input image and high-quality edits. We run extensive experiments and compare our method with state-of-the-art inversion methods. Qualitative metrics and visual comparisons show significant improvements.
\end{abstract}

\begin{IEEEkeywords}
GAN inversion, Image editing, Generative Adversarial Networks.
\end{IEEEkeywords}}

% make the title area
\maketitle

\IEEEdisplaynontitleabstractindextext
\IEEEpeerreviewmaketitle

\section{Introduction}
\label{sec:intro}

Generative Adversarial Networks (GANs) significantly advanced the image generation field by synthesizing high-quality images ~\cite{karras2019style, karras2020analyzing, yu2021dual, dundar2023fine, dundar2023progressive}.
Those GAN models that are trained on large-scale image datasets for image generation tasks are also shown to achieve realistic image edits even when they were not trained for editing objectives.
They achieve that by the emergence of semantically interpretable latent spaces in an unsupervised way. 

\begin{figure*}
\centering
\includegraphics[width=1.0\textwidth]{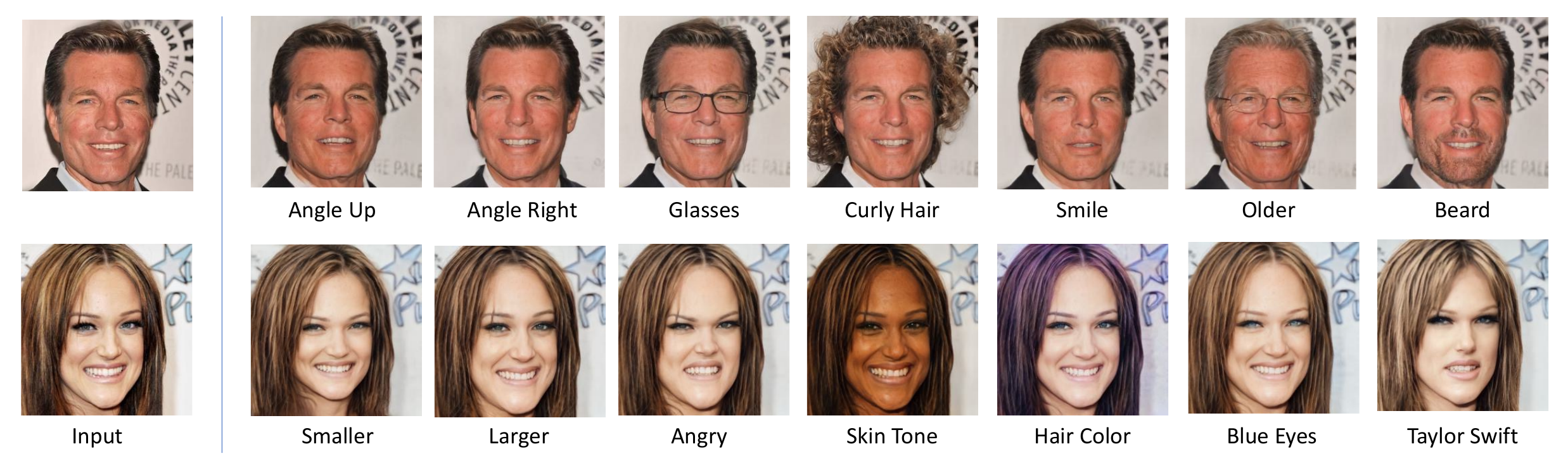}
\caption{Image editing capabilities with GAN inversion. Input images are inverted with our framework and edits are performed with the edit directions found by InterfaceGAN \cite{shen2020interpreting}, GanSpace \cite{harkonen2020ganspace}, and StyleClip \cite{patashnik2021styleclip}.}
\label{fig:teaser}
\end{figure*}

There has been extensive research in exploring latent directions of GANs to edit images \cite{shen2020interpreting,voynov2020unsupervised,harkonen2020ganspace,shen2021closed,wu2021stylespace,abdal2021styleflow, wu2021stylespace, pan2023drag}. Edits span from changing the pose of objects to changing the hairstyle of a person to make someone look more like Taylor Swift \cite{patashnik2021styleclip}. A few of the possible edits are shown in Fig. \ref{fig:teaser}, where input images are inverted with our proposed framework.
Such capabilities are beyond what can be achieved by the image translation methods that require image sets for each attribute for their training \cite{starganv2, li2021image,yang2021l2m, dalva2022vecgan, dalva2023image}. 
On the other hand, for real image editing with pretrained GAN models, an image has to be inverted to GAN's latent space and that is challenging. 

The challenge exists because of the trade-off between the image reconstruction fidelity and editing quality. 
For example, in StyleGAN \cite{karras2019style}, when an image is inverted to its $W^+$ space, it is compressed to $18\times512$ dimension. While it is editable, the reconstruction loses many image details due to the information bottleneck \cite{richardson2021encoding, tov2021designing, yildirim2023diverse}.
The reconstruction can be improved by letting higher dimensional latent features pass to the GAN generator, but then edit quality decreases if those features become inconsistent with the edits \cite{wang2022high, pehlivan2023styleres}.

To tackle this trade-off, our previous work, StyleRes \cite{pehlivan2023styleres}, proposes a framework that can learn residual features in higher-rate latent codes that are missing in the reconstruction of encoded features.
However, those high-rate latent codes need to adapt to the image edits. 
For example, if high-rate codes carry information about the earring of a person and the image is edited to change the pose of the person, then those codes should be carried to the generator accordingly to appear at the correct place. 
Otherwise, it will cause a ghosting effect. 
Because of that reason, StyleRes proposes to train the model with random edits and with cycle consistency guidance such that when an edit is applied and reverted back, the original image is reconstructed. Via this guidance, it learns a module to transform these higher-rate latent codes for image editing.
However, the learned module is a convolutional neural network and has limited power in transforming the residuals correctly for edits.

In this work, our main contribution is a pipeline that can learn the residuals for high-fidelity image reconstruction and adopt them correctly for high-quality image edits. 
Our framework achieves that by estimating flows between the original and edited image features and warping the residual features according to the flow.
We take advantage of an unsupervised flow estimation network \cite{truong2021warp} to train our framework in a novel way. 
This framework achieves significantly better results under extreme edit transformations while our architecture is single-stage and efficient.

Our contributions are as follows 
\begin{itemize}
    \item We propose a novel inversion pipeline that achieves high-quality image edits by predicting a flow between the original and edited image and learning to warp the high-rate features based on the predicted flows.
    \item We guide the flow prediction via an unsupervised pretrained flow estimation network. The flow prediction takes StyleGAN intermediate features as inputs to achieve efficiency.
    \item We show that our framework can work with different pretrained StyleGAN inversion networks and improve all of them with large margins. 
    \item Our extensive experiments show the effectiveness of our framework and achieve significant improvements over state-of-the-art for both reconstruction and real image attribute manipulations.
    \item We provide a comparison of our method, which is built for StyleGAN inversion and an inversion model built for 3D-aware GAN models. We discuss their different advantages and provide an analysis.

\end{itemize}

\section{Related Work}

Image editing requires high-fidelity reconstruction of input images and being able to only change the selected attributes of images while preserving others.  
Previously, separate image translation models were trained for each attribute change \cite{zhu2017unpaired, dundar2020panoptic}. Later, it is shown that a single image translation network can be trained to achieve multiple edits \cite{starganv2, yang2021l2m, dalva2022vecgan, dalva2023image}. 
These models require labeled datasets for each attribute. For example, they require a set of images that have people with eyeglasses and another set without glasses to learn the translation of removing or adding eyeglasses.
Because of that reason, those image translation methods offer a limited amount of attribute changes. 
For example, VecGAN \cite{dalva2022vecgan} and L2MGAN \cite{yang2021l2m} showcase 5-6 different edits on the CelebA dataset (hair color, smile, gender, age, bangs) \cite{celeba}.

One important discovery for image editing research, especially after StyleGAN models \cite{karras2020analyzing}, is that well-trained GAN models organize their latent space in a semantically interpretable way and have the power to edit images. 
Since then, numerous edits have been discovered in unsupervised ways on StyleGAN latent space \cite{harkonen2020ganspace, abdal2021styleflow, voynov2020unsupervised,shen2021closed,wu2021stylespace}.  
Especially, with the  CLIP guidance \cite{patashnik2021styleclip} for the discovery of directions with text-based prompts, it is shown that images can be edited even to make someone look like a specific celebrity or the style of an image can be changed to gothic style with text prompts.  
To enjoy such editing capabilities of StyleGAN, a real image should be first converted to StyleGAN's latent space, which is referred to as GAN inversion and is the topic of this paper.

GAN inversion is initially explored by latent optimization such that latent codes are optimized by back-propagating the reconstruction loss between the generated and target images \cite{creswell2018inverting, abdal2019image2stylegan, abdal2020image2stylegan++, karras2020analyzing, roich2022pivotal}. 
These methods are slow since for each image, latent codes need to be optimized.
Additionally, even though the reconstruction loss may be minimized, there is no guarantee that the image is inverted to the natural latent space of GANs. 
For image editing to be possible, images need to be inverted to the natural GAN space where directions are interpretable. 
Because of that reason and to achieve high speed, image encoders are trained for the inversion task \cite{zhu2020domain, richardson2021encoding,  tov2021designing}.
These encoders leverage the knowledge of training sets while projecting images and can also be trained with additional objectives than image reconstructions, such as discriminators that compare the generated and real images as well as discriminators in latent space to achieve inversion to GAN's natural latent space.
They are also faster. 
However, they are not very accurate in the reconstruction of input image details. 
This is mainly because these models project images into $W^+$ space of StyleGAN, which has $18\times512$ dimension. 
This reduction results in information loss and input images cannot be generated with high fidelity.

\begin{figure*}[t]
\centering
\includegraphics[width=\linewidth]{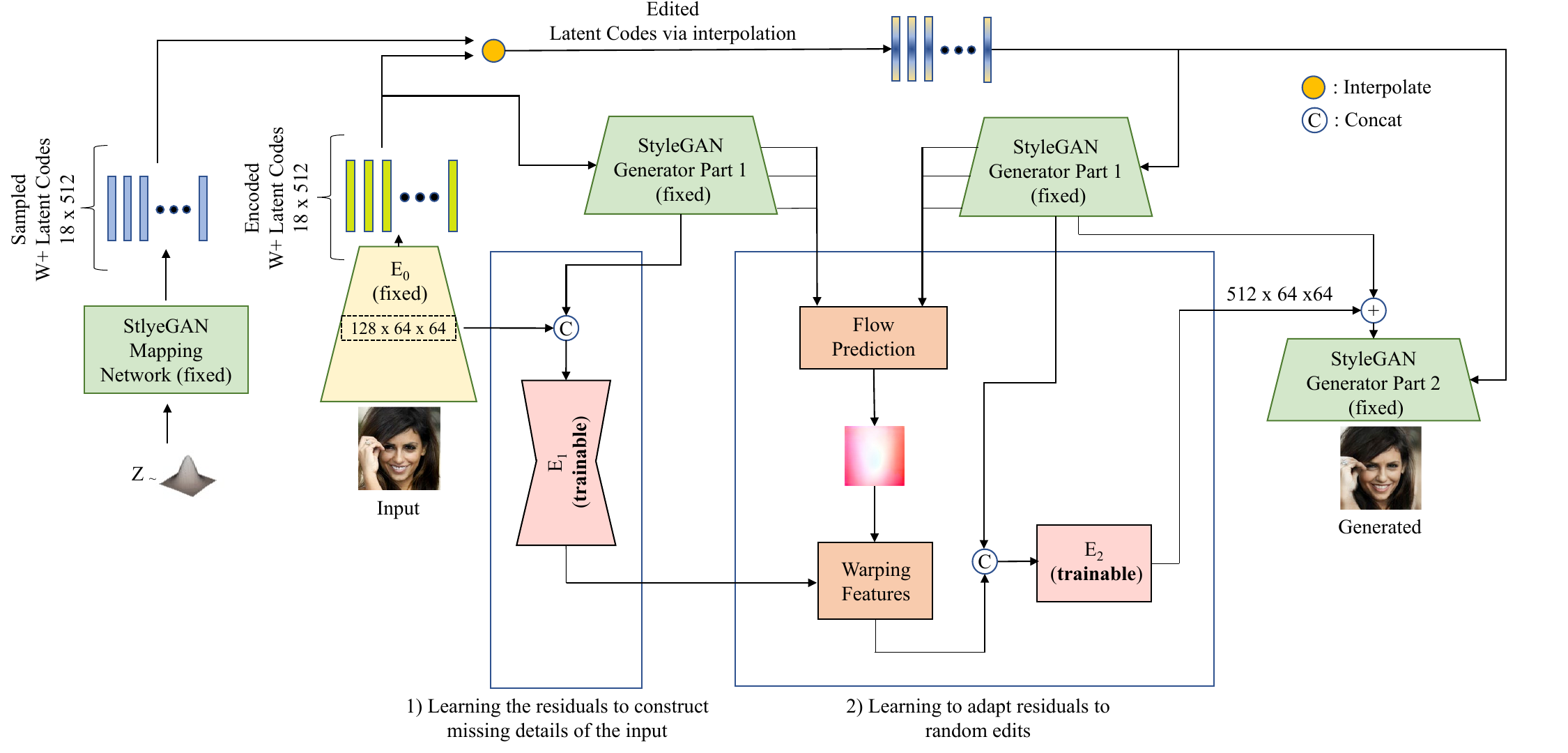}
\caption{WarpRes proposes a pipeline that predicts low-rate latent codes via a pretrained $E_0$ and high-rate latent codes via $E_1$. The low-rate latent codes, $W^+$, are used for generation via StyleGAN and provide editing capabilities by linear operations on the $W^+$ space.
High-rate latent codes are needed for high-fidelity reconstruction, however, because they will be added to the StyleGAN generator at the feature level, there is a need for a mechanism to transform them based on the edits.
Because of that reason, we warp high-rate latent codes via the flow predictions based on the edits. 
Flows are predicted on StyleGAN mid-level feature generations of original and edited, $W^+$.
After warping the high-rate latent codes, we process them via $E_2$ for final refining of features before adding them to StyleGAN mid-level features.
Same as StyleRes \cite{pehlivan2023styleres}, during training, latent codes are edited by interpolating encoded W+'s with randomly generated ones by StyleGAN's mapping network. During inference, they are edited with semantically meaningful directions discovered by methods such as InterfaceGAN and GANSpace.
}
\label{fig:architecture}
\end{figure*}

To achieve high-fidelity, two-stage feed-forward encoders are explored \cite{alaluf2021restyle, alaluf2022hyperstyle, wang2022high, pehlivan2023styleres}. In these methods, first images are inverted to $W^+$ space with a pretrained encoder \cite{richardson2021encoding}, and missing features that were lost via the information bottleneck are recovered by letting higher-rate features from a second-stage encoder to the StyleGAN generator \cite{alaluf2021restyle, alaluf2022hyperstyle, wang2022high}. The challenge in these methods is that those high-rate features need to adapt to the image edits because editing is the goal of the GAN inversion in the first place. Therefore, those high-rate features need to be transformed based on the edited features.
StyleRes \cite{pehlivan2023styleres} is the most successful method in adapting the high-rate features to image edits via its network design and cycle-consistency-based loss objectives.
However, StyleRes also struggles when edits require large transformations such as in pose edits.
In this work, we design a novel framework that includes flow estimation between the original and edited features and warp the high-rate features based on the flow estimates.

\section{Method}

\subsection{WarpRes Architecture}
\label{sec:arch}

Our architecture is shown in Fig. \ref{fig:architecture} and built on our previous conference work, StyleRes \cite{pehlivan2023styleres}.
Same as StyleRes, we utilize an encoder $E_0$ that can embed images into $W+$ latent space and a StyleGAN generator $G$ \cite{karras2020analyzing}.
For the pretrained encoder, similar to previous works \cite{wang2022high, pehlivan2023styleres}, we use e4e \cite{tov2021designing} but in our experiments later, we show that our framework can use and improve over the other pretrained encoders as well.

Our architecture first extracts high-level features from the encoder since with only low-rate latent codes, the reconstruction is poor. 
Both the high and low rate features are extracted from the pretrained encoder, $E_0$; $F_0, W^+ = E_0(x)$.
After that, both StyleRes and WarpRes utilize a second encoder, $E_1$ whose goal is to detect the missing features of an input image by comparing the extracted features, $F_0$ and StyleGAN generated features, $F_g$ in the mid-level.
We extract $F_g$ from StyleGAN generation at $64\times64$ dimension.

\begin{equation}
\label{eq:E1}
 F_a = E_1( F_0, F_g)
\end{equation}

$E_1$ is expected to compare $F_0$ and $F_g$ to detect the missing features and output the residual features that are needed for high fidelity.
$E_1$ does not receive any edited image features and so cannot adopt features based on edits.
After that, WarpRes architecture differs from StyleRes. 
StyleRes directly sends features to the second encoder, $E_2$ which has the goal of transforming the residual features extracted by $E_1$ based on the edited features coming from StyleGAN intermediate layers, $F_e$.

During training, edits are simulated by taking random directions from encoded $W^+$.
The random directions are produced as follows: 
A $z$ vector is sampled from the normal distribution and via StyleGAN's mapping network,  $W^+_r$ is outputted, $M$; $W^+_r=M(z)$.
Next, we take a step towards $W^+_r$ to obtain a mixed style code $W^+_\alpha$. 

\begin{equation}
\label{eq:alpha}
 W^+_\alpha= W^+ + \alpha ({W^+_r - W^+})
 \end{equation}

where  $\alpha$ controls the degree of the edit.
If $\alpha$ is set to $0$, that corresponds to no edit. 
We either set  $\alpha$ to $0$ or a value in the range of $(0.4, 0.5)$ to simulate an edit.
We extract again the StyleGAN generated mid-features, $F_e$ from $64\times64$ but this time features are generated with $W^+_\alpha$.
The edits output different features, for example, $F_g$ may be features generated for a person at one pose and $F_e$  may be from another pose. 
In WarpRes, we feed $F_e$ and $F_g$ to a flow prediction network together with StyleGAN generator's previous features one from $16\times16$ and another from $32\times32$. 
Our goal is to predict the transformation of features based on edits so that we can warp the residual features, $F_a$ accordingly.
We guide this flow prediction via a pretrained flow network \cite{truong2021warp}. 
This pretrained flow network that we use for guidance expects RGB images, so for training, we have to generate images for $W^+$ and $W^+_\alpha$.
We estimate the flow on generated images that are in $256\times256$ dimension and downsample and divide the flow prediction by $4$ for $64\times64$ dimension.
These predictions are used as ground truth when we guide the flow prediction network that operates on StyleGAN-generated features and estimates a flow in the $64\times64$ dimension.
The flow predictions are used to warp the residual features $F_a$ to obtain $F_{wa}$.
Next, we feed $F_{wa}$ and $F_e$ to the last encoder for the warped features to additionally tune before passing to the StyleGAN generator as below:

\begin{equation}
\label{eq:E2}
F = E_2(F_{wa}, F_e)
\end{equation}

Finally, $F$ and  $F_e$ are summed, and given as input to the next convolutional layers in the generator.  

\newcommand{\interpfigt}[1]{\includegraphics[trim=0 0 0cm 0, clip, width=2.6cm]{#1}}

\begin{table*}[]
\centering
\caption{Ablation study results of reconstruction and editing scores on CelebA-HQ dataset. For reconstruction, we report FID and LPIPS scores. For editing, we report FID metrics for smile and pose addition (+) and removal (-). We also report Id scores to measure if identity is preserved during edits.
The higher the Id score, the better, and for the other metrics, the lower value is better.}
\begin{tabular}{|l||l|l||l|l|l|l||l|l|l|l|}
\hline
 & \multicolumn{2}{c||}{Reconstruction} & \multicolumn{4}{c||}{Editing - Smile} & \multicolumn{4}{c|}{Editing - Pose}  \\
\hline
Method &  FID   & LPIPS &  FID(+) & FID(-)  &  Id(+) & Id(-) & FID(+) & FID(-) & Id(+) & Id(-) \\
\hline
StyleRes \cite{pehlivan2023styleres}  & 7.04  & 0.09 & 23.52 & 21.80 & 0.49 & 0.44 & 11.31 & 10.73 & 0.58 & 0.59  \\  
Pretrained flow  &  7.55  &   0.09  &  23.41 & 22.41 & 0.54 & 0.51 & 10.66 & 10.40 & 0.67 & 0.68  \\   
Trained flow  w/o flow loss & 7.24 & 0.08 & 23.29 & 24.54 & 0.51 & 0.42 & 16.38 & 15.87 & 0.56 & 0.56 \\
Trained flow  & 6.59 & 0.08 &  22.46 & 21.52 & 0.57 & 0.59 & 10.40 & 10.04 & 0.74 &  0.74 \\
\hline
 pSp \cite{richardson2021encoding} &  23.86 & 0.17 & 32.47 & 34.00 & 0.45 & 0.44 & 22.95 & 22.13 & 0.50 & 0.51 \\  
 pSp \cite{richardson2021encoding}+ StyleRes & 7.12  & 0.08 & 21.11 & 20.80 & 0.59 & 0.58  & 12.71 & 11.85 & 0.58 & 0.60 \\ %%
pSp \cite{richardson2021encoding} + WarpRes   & 6.60 & 0.08 &  20.70 & 21.76 & 0.66 & 0.69 & 10.67 & 9.62 & 0.72 & 0.75 \\ 
\hline
e4e \cite{tov2021designing} &  30.22 &  0.21 & 38.58 & 39.68 & 0.38 & 0.32 & 26.92 & 26.83 & 0.46 & 0.46  \\
e4e \cite{tov2021designing} + StyleRes  & 7.04  & 0.09 & 23.52 & 21.80 & 0.49 & 0.44 & 11.31 & 10.73 & 0.58 & 0.59 \\   
e4e \cite{tov2021designing} + WarpRes  & 6.59 & 0.08 &  22.46 & 21.52 & 0.57 & 0.59 & 10.40 & 10.04 & 0.74 &  0.74 \\ 
\hline
StyleTransformer \cite{Hu_2022_CVPR}& 21.82 & 0.17 & 34.32 & 34.61 & 0.45 & 0.39 & 20.26 & 21.06 & 0.52 & 0.54 \\
StyleTransformer \cite{Hu_2022_CVPR} + StyleRes   & 7.63  &   0.08  & 24.09 &  22.60 & 0.54 & 0.47 & 13.61 & 13.90 & 0.57 & 0.60 \\ %%
StyleTransformer \cite{Hu_2022_CVPR} + WarpRes   & 6.85 & 0.08 & 21.82 & 22.08 & 0.62 & 0.58 & 12.00 & 12.53 & 0.67 & 0.71 \\ 
\hline
\end{tabular}
%}
\label{table:ablation}
\end{table*}

Architecture details are as follows: Our model uses a series of residual layers, which are visualized in Fig. \ref{fig:architecture}. Encoder $E_1$ takes the concatenated features $F_0$ and $F_{g}$, which has a spatial dimension of $640\times64\times64$.
$E_1$ network has an encoder-decoder architecture with 3 downsampling of ResNet blocks and 3 upsampling of ResNet blocks.
ResNet blocks have two convolutional layers with a Leaky Relu layer in between. 
There is a skip connection from the input to the output in the form of summation.
The channel size stays as $128$ at each layer and filters are $3\times3$ in convolutional layers.
There are skip connections from each residual block of the encoder to its corresponding block from the decoder which operates on the same spatial dimension.
The output of E1, $F_a$, has the  spatial resolution of  $512\times64\times64$. 
This feature is warped using the flow prediction network to get $F_{wa}$. 

Then, our second encoder $E_2$ is fed with the features $F_{wa}$ and $F_e$. Before concatenating these features, we reduce their channel size by half via the $3\times3$ convolutional layer. This network is simpler without downsampling or upsampling operations. It is a flat architecture with 4 blocks of residual convolutional layers with a channel size of $128$ and with $3\times3$ filters and a padding size of 1. 
The last channel size is $512$ and the model outputs $512\times64\times64$ dimensional features, using residual layers. 

Normally, the flow prediction network uses VGG features \cite{simonyan2014very} of RGB images extracted at $16\times16$, $32\times32$, and $64\times64$ resolutions. We want to avoid generating an RGB image and extract its features because of performance issues. Instead, we collect StyleGAN features at the same resolutions while we extract $F_a$ and $F_e$. The channel dimensions of StyleGAN and VGG features do not match, so we use $3\times3$ convolutions to match them. The weights of the modified flow network are initialized with the weights of the pretrained flow network \cite{truong2021warp}. We omit the details of the flow architecture for brevity,  same architecture is used from  \cite{truong2021warp}. We fine-tune the flow estimation module together with $E_1$ and $E_2$ with the losses described in the next section.

\subsection{Training Phases}
\label{sec:training}

We train our model with the no editing and cycle translation paths proposed by StyleRes \cite{pehlivan2023styleres}.

\textbf{No Editing Path.} In this path, $\alpha$ from Eq. \ref{eq:alpha} is set to $0$ and so the target image is the same as the input image. This path is set to train the network to reconstruct high-fidelity details but does not teach the model how to adapt to edits.

\textbf{Cycle Translation Path.} This path uses the random edit directions as given in Eq. \ref{eq:alpha}.
With the random edits, we generate an output image $x_i'$. $x_i'$ generation can be guided with adversarial losses to be realistic but then it will not have any incentive to carry realistic details from the input image.
For that, we set a cycle translation path which includes encoding $x_i'$ and taking the step back in the editing direction to reverse the edit.
With that, the generator reconstructs $x''$, which is supposed to match the input image $x$.

\subsection{Training Objectives}
\label{sec:losses}

\textbf{Reconstruction Losses.} 
We use the output of no editing path, $x'$, and the output of cycle translation path, $x''$ in our reconstruction-based losses. 
Both of those outputs are supposed to match the input image. 
To supervise this behavior, we use  $L_2$ loss, perceptual loss, and identity loss between the input and output images:

\begin{equation}
    \begin{split}
        \mathcal{L}_{rec-l2} = ||x' - x||_2 + 
        ||x'' - x||_2
    \end{split}
    \label{eqn:rec_loss}
\end{equation}

where $x$ is both the input and target image.
We use perceptual losses from VGG ($\Phi$) at different feature layers ($j$) between these images from the loss objective as given in Eq. \ref{eq:percep}. 

\begin{equation}
\label{eq:percep}
\small
    \mathcal{L}_{rec-p} = ||\Phi_{j}(x') - \Phi_{j}(x) ||_2 + ||\Phi_{j}(x'') - \Phi_{j}(x) ||_2
\end{equation}

Similar to previous works \cite{alaluf2022hyperstyle}, we use identity loss with a pretrained network $A$. $A$ is an ArcFace model \cite{deng2019arcface} when training on the face domain and can compare the identity of two face images. It is a domain-specific ResNet-50 model \cite{tov2021designing} for our training on car class. 
\begin{equation}
    L_{rec-id} = (1 - \langle A(x),A(x')\rangle)+(1 - \langle A(x),A(x'')\rangle)
\end{equation}

\textbf{Adversarial Losses.} When cycle consistency is used for training, the intermediate image generation may output something noisy and still can achieve cycle consistency.
Because of that reason, it is important to guide the intermediate image generation, $x_i'$, to be realistic. We use adversarial losses to guide the network's output to realistic images whether there is an edit ($x_i'$) or no edit ($x'$).
We load the pretrained discriminator from StyleGAN training, $D$, and train the discriminator together with the encoders.

\begin{equation}
 \begin{split}
    L_{adv} = 2\log{D(x)} + \log{(1-D(x'))} \\ + \log{(1-D(x'_i))}
    \end{split}
\end{equation}

\textbf{Feature Regularizer.} We want the main generation to be achieved with low-rate latent codes and only allow high-rate features that are needed for high-quality reconstruction even though high-rate features alone can achieve perfect reconstruction. To prevent high-rate features from coding everything needed for the reconstruction, same as StyleRes \cite{pehlivan2023styleres}, we regularize the residual features to be small:
\begin{equation}
    L_{F} = \sum_{F \in \phi} \|F\|_1
\end{equation}

\textbf{Flow Estimation Loss.} Different than StyleRes \cite{pehlivan2023styleres}, we guide the flow prediction network with pseudo-ground-truth flow predictions. We obtain the pseudo ground-truth predictions by running a pretrained flow network that was trained in an unsupervised way \cite{truong2021warp}. This pretrained network predicts the flow from edited and unedited images generated by StyleGAN. We use $L_1$ loss between the flow prediction and pseudo ground-truths to obtain the $L_{flow}$ objective.

\textbf{Full Objective.}
We use the overall objectives given below. 

\begin{equation}
    \begin{split}
        \underset{E_1,E_2}{\min} \underset{D}{\max} \lambda_{a}\mathcal{L}_{adv} +  \lambda_{r1} \mathcal{L}_{rec-l2} + \lambda_{r2} \mathcal{L}_{rec-p}\\
        +\lambda_{r3} \mathcal{L}_{rec-id}+
        \lambda_{f} \mathcal{L}_{F} +  \lambda_{fl} \mathcal{L}_{flow}
    \end{split}
    \label{eqn:full_loss}
\end{equation}

When we choose the \emph{no editing path}, we set $\lambda_{r1}=1.0$, $\lambda_{r2}=0.001$, $\lambda_{r3}=0.1$ for the face and $\lambda_{r3}=0.5$ for the car dataset. When we choose the \emph{cycle translation path}, we set $\lambda_{r1}=0.0$, meaning we do not use cycle consistency at the pixel level, $\lambda_{r2}=0.0001$,  $\lambda_{r3}=0.01$ for the face and $\lambda_{r3}=0.05$ for the car dataset. At both paths, we set $\lambda_{a}=0.1$. The regularizer coefficient is set to $\lambda_{f}=3.0$ for both datasets.  
The network is trained with Adam optimizer, with a learning rate equal to $0.0001$. We halved the learning rate at iterations 5000, 10000, and 15000. 

\begin{figure*}
\centering
\scalebox{0.71}{
\addtolength{\tabcolsep}{-5pt}   
\begin{tabular}{ccccccccc}

\rotatebox{90}{~~~~~~~1.  Pose (+)} &
\interpfigt{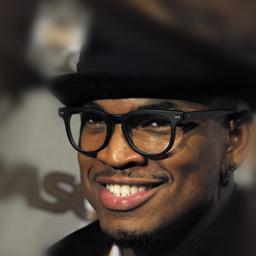} &
\interpfigt{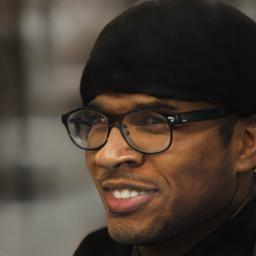} &
\interpfigt{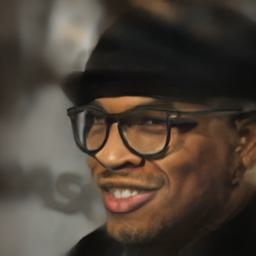} &
\interpfigt{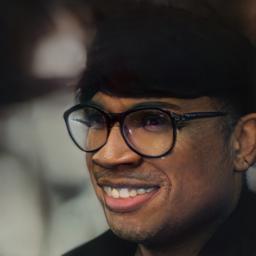} &
\interpfigt{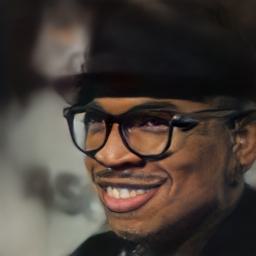} &
\interpfigt{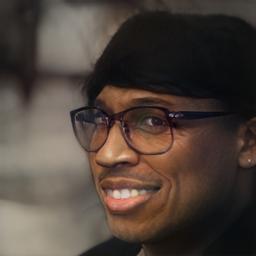} &
\interpfigt{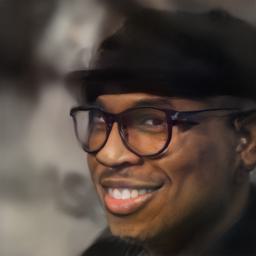} &
\\
\rotatebox{90}{~~~~~~~2.  Pose (+)} &
\interpfigt{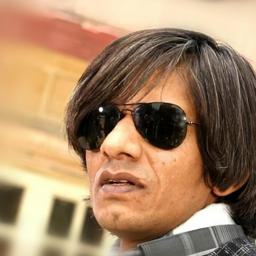} &
\interpfigt{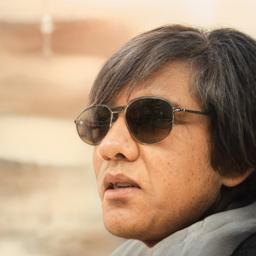} &
\interpfigt{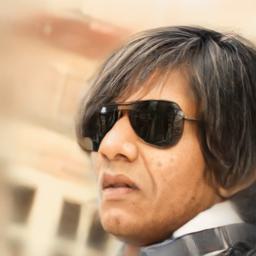} &
\interpfigt{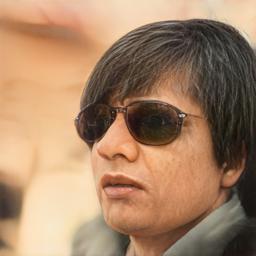} &
\interpfigt{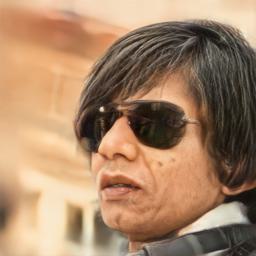} &
\interpfigt{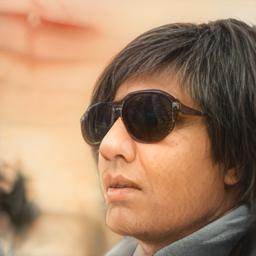} &
\interpfigt{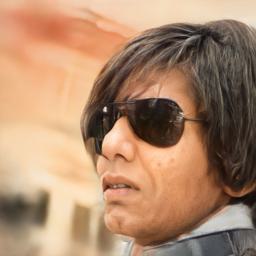} &
\\
\rotatebox{90}{~~~~~~~3.  Age (+)} &
\interpfigt{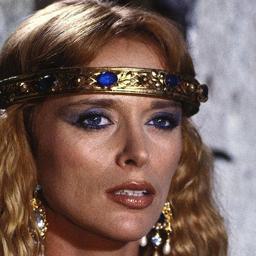} &
\interpfigt{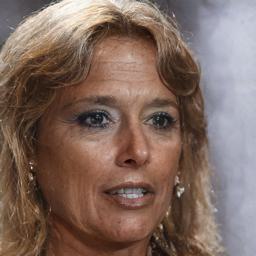} &
\interpfigt{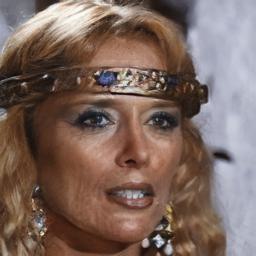} &
\interpfigt{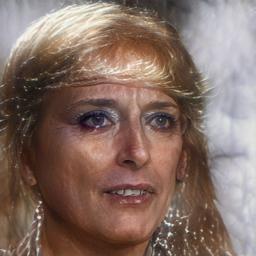} &
\interpfigt{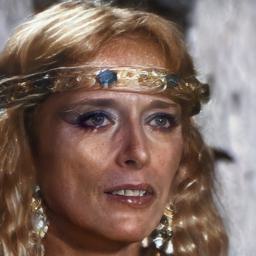} &
\interpfigt{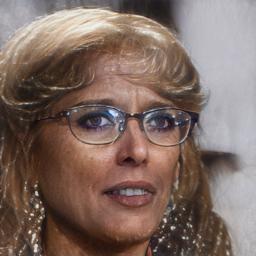} &
\interpfigt{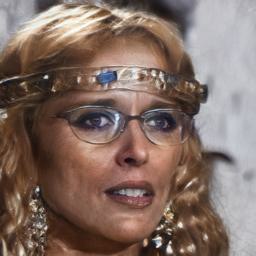} &
\\
\rotatebox{90}{~~~~~~~4.  Age (+)} &
\interpfigt{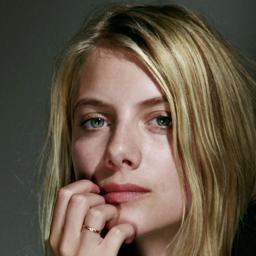} &
\interpfigt{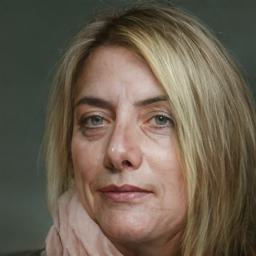} &
\interpfigt{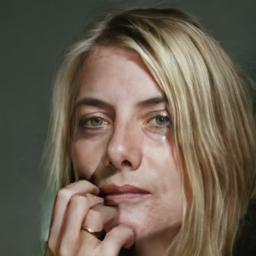} &
\interpfigt{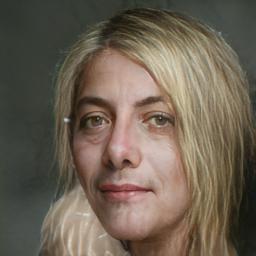} &
\interpfigt{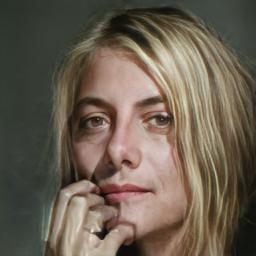} &
\interpfigt{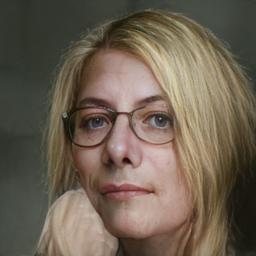} &
\interpfigt{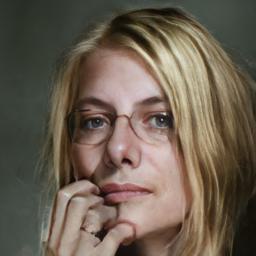} &
\\
\rotatebox{90}{~~~~~~~5.  Age (+)} &
\interpfigt{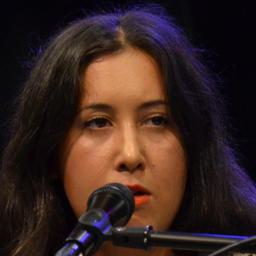} &
\interpfigt{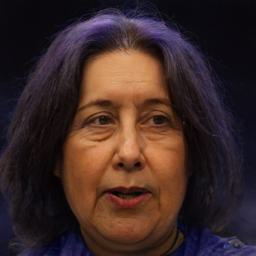} &
\interpfigt{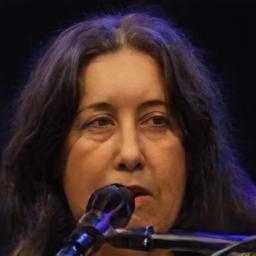} &
\interpfigt{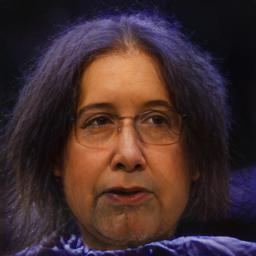} &
\interpfigt{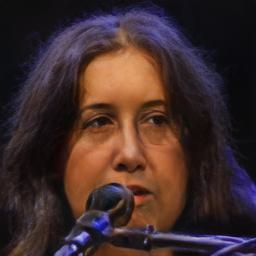} &
\interpfigt{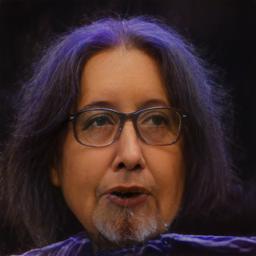} &
\interpfigt{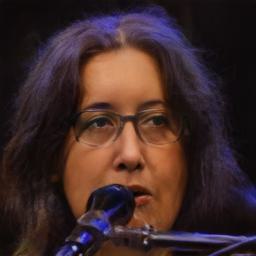} &
\\
\rotatebox{90}{~~~~~~~6.  Age (+)} &
\interpfigt{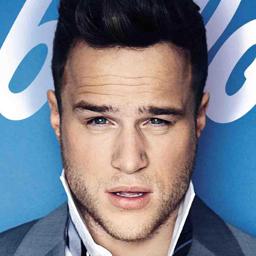} &
\interpfigt{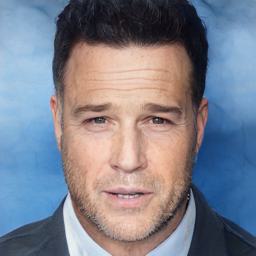} &
\interpfigt{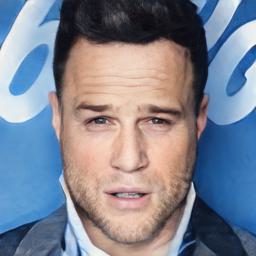} &
\interpfigt{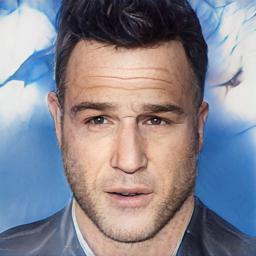} &
\interpfigt{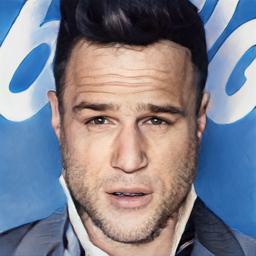} &
\interpfigt{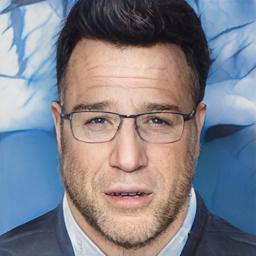} &
\interpfigt{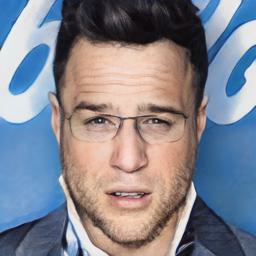} &
\\
& Input &   e4e  & e4e + Warpres  & pSp   & pSp + Warpres & StyleTransformer & ST + WarpRes \\
\end{tabular}
}
\caption{Qualitative results of different base encoders and encoders with WarpRes. ST + WarpRes refers to the WarpRes model with StyleTransformer base architecture.
Our model is able to work with different encoders. Encoders miss many image details and reconstruct the images with low fidelity while WarpRes is able to provide the details and correct the image details. For example, in the first two rows, encoders alone generate images with different eye-glasses while WarpRes is able to correct them and additionally add wrinkles correctly which helps to preserve the identity.}
\label{fig:results_bases}
\end{figure*}

\begin{table*}[t]
\centering
\caption{Quantitative results of reconstruction and editing on CelebA-HQ dataset. For reconstruction, we report FID, SSIM, and LPIPS scores. For editing, we report FID metrics for smile addition (+) and removal (-).
We also report Id scores to measure if identity is preserved during edits.
The higher the Id and SSIM scores, the better, and for FID and LPIPS metrics, the lower value is better.}
\begin{tabular}{|l|l||l|l|l||l|l|l|l||l|l|l|l|}
\hline
& Runtime & \multicolumn{3}{c||}{Reconstruction} & \multicolumn{4}{c||}{Editing - Smile} & \multicolumn{4}{c|}{Editing - Pose}  \\
\hline
\textbf{Method} & (sec) & FID  & SSIM  & LPIPS &  FID(+) & FID(-)  &  Id(+) & Id(-) & FID(+) & FID(-) & Id(+) & Id(-) \\
\hline
pSp \cite{richardson2021encoding} &  0.0337 &  23.86 & 0.75 & 0.17 & 32.47 & 34.0 & 0.45 & 0.44 & 22.95 & 22.13 & 0.50 & 0.51 \\  
e4e \cite{tov2021designing} & 0.0303 & 30.22 & 0.71 & 0.21 & 38.58 & 39.68 & 0.38 & 0.32 & 26.92 & 26.83 & 0.46 & 0.46 \\
ReStyle \cite{alaluf2021restyle} & 0.3715 & 24.82 & 0.73 & 0.20 & 30.35 & 33.69 & 0.42 & 0.42 & 21.65 & 22.43 & 0.49 & 0.49 \\
HyperStyle \cite{alaluf2022hyperstyle}& 0.4496 &16.08 & 0.83 &  0.11  &  26.43 & 25.26 & 0.56 & 0.57 & 15.59 & 15.83 & 0.68 & 0.69 \\
HFGI \cite{wang2022high} &  0.0551 &12.17 & 0.85 & 0.13 & 25.22 & 27.10 & 0.45 & 0.39 & 15.37 & 15.97 & 0.54 & 0.53 \\
StyleTransformer \cite{Hu_2022_CVPR} &  0.0297 &21.82 & 0.75 & 0.17 & 34.32 & 34.61 & 0.45 & 0.39 & 20.26 & 21.06 & 0.52 & 0.54 \\
FeatureStyle \cite{xuyao2022} & 0.3780 & 11.33 & 0.90 & 0.10 & 27.20 & 26.15 & 0.51 & 0.45 & 30.00 & 24.24 & 0.62 & 0.63 \\
StyleRes \cite{pehlivan2023styleres}& 0.0429 & 7.04 & 0.90 & 0.09 & 23.52 & 21.80 & 0.49 & 0.44 & 11.31 & 10.73 & 0.58 & 0.59  \\   
\hline
 WarpRes &  0.1254 & \textbf{6.59} & \textbf{0.91} & \textbf{0.08} &  \textbf{22.46} & \textbf{21.52} & \textbf{0.57} & \textbf{0.59} & \textbf{10.40} & \textbf{10.04} & \textbf{0.74} &  \textbf{0.74} \\
\hline
\end{tabular}
%}
\label{table:results_face}
\end{table*}

\begin{table*}[t]
\centering
\caption{Quantitative results of reconstruction and editing on the Stanford Cars Dataset. For reconstruction, we report FID, SSIM, and LPIPS scores. For editing, we report FID metrics for grass addition, color, and viewpoint change.}
\begin{tabular}{|l|l|l|l||l|l|l|l|c|c|}
\hline
& \multicolumn{3}{c||}{Reconstruction} & \multicolumn{3}{c|}{Editing - FIDs}  \\
\hline
\textbf{Method} & FID  & SSIM  & LPIPS &  Grass & Color & Viewpoint \\
\hline
e4e \cite{tov2021designing} & 14.04 & 0.50 & 0.32 & 18.02 & 29.79 & 16.72 \\
ReStyle \cite{alaluf2021restyle}& 13.38 & 0.57 & 0.30 & 16.01 & 21.34 & 14.80 \\
HyperStyle \cite{alaluf2022hyperstyle}& 11.64 & 0.63 & 0.28 &  17.13 & 26.30 & 13.56\\
HFGI \cite{wang2022high} & 9.41 & 0.83  & 0.16 & 14.84 & 26.65 & 21.80 \\
StyleTransformer \cite{Hu_2022_CVPR} & 14.01 & 0.57 & 0.28 & 19.47 & 19.94 & 16.55 \\
StyleRes  \cite{pehlivan2023styleres} &  7.60 &  0.83 &  0.14 &  10.64 & 18.86 & 12.62  \\ 
\hline
WarpRes & \textbf{5.94} & \textbf{0.87} & \textbf{0.12} & \textbf{8.51} & \textbf{17.49} & \textbf{10.85} \\ % HF -1: 12.63 (HF 1 is reported)
\hline
\end{tabular}
\label{table:results_car}
\end{table*}

\section{Experiments}

\textbf{Set-up.} We run experiments on the human face and car datasets. For the human face domain, we train the model on the FFHQ \cite{karras2019style} dataset and evaluate it on the CelebA-HQ \cite{karras2017progressive} dataset same as previous works.
For the car domain, we use Stanford Cars \cite{krause20133d} with their train evaluation splits.

\textbf{Evaluation.}
We report metrics for reconstruction and editing qualities as well as run-time comparisons.
Reconstruction refers to generating original input images without any edits. 
For the reconstruction task, we report Frechet Inception Distance (FID) metric \cite{heusel2017gans}, Learned Perceptual Image Patch Similarity (LPIPS) \cite{zhang2018unreasonable} and Structural Similarity Index (SSIM).
FID looks at realism by comparing generated and original image distributions in the Inception network feature level \cite{szegedy2015going}. 
LPIPS compares the output image with the ground truth again at the feature level by utilizing a pretrained VGG network \cite{simonyan2014very}. 
SSIM compares the input and output images at pixel level. 
We are also interested in editing qualities and for the CelebA-HQ dataset, we look at smile and pose editings discovered by InterfaceGAN \cite{shen2020interpreting} given their challenges, and for the Stanford cars dataset, we change grass, color, and pose attribute of images. We use these edits that are found by the GanSpace method \cite{harkonen2020ganspace}.
We report FID metrics for smile by taking advantage of the CelebA attribute labels as follows: We use the test set that are labeled as not smiling and add smile to them, which provides us the generated image set. For the real image set, we take the test set that is labeled as smiling. 
FID is calculated between them, this way the image quality and also the attribute addition qualities are measured by FIDs. The same set-up is also used for smile removal. 
On the other hand, for pose, we use the entire test set.
We do not have ground-truth images to report LPIPS, however, we report identity similarity (Id) scores between the original and edited images. 
Edits are not supposed to change the identity of a person, therefore, we expect a high Id score. 
We utilize the CurricularFace model \cite{huang2020curricularface} to measure the identity scores.  
On the Stanford cars dataset, we calculate the FIDs between the edited and original images.

We also report runtimes of different methods.
The running times are obtained by averaging the reconstruction times of 2000 samples with batch size equal to 1 on a single NVIDIA GeForce GTX 1080 Ti GPU.
Reconstruction times include encoding the image and generation with StyleGAN.

\textbf{Baselines.} Our comparisons include single-stage encoders that output low-rate latent codes such as pSp \cite{richardson2021encoding}, e4e \cite{tov2021designing}, and StyleTransformer \cite{Hu_2022_CVPR} as well as frameworks that also predicts higher-rate latent codes via a second stage such as  ReStyle \cite{alaluf2021restyle}, HyperStyle \cite{alaluf2022hyperstyle}, HFGI \cite{wang2022high},  FeatureStyle \cite{xuyao2022}, and StyleRes \cite{pehlivan2023styleres}.
We use the released models from authors for CelebA dataset.
For the car dataset, we compare it with the models that provide released models.

\textbf{Ablation Study.}  We start our experiments with an ablation study as presented in Table \ref{table:ablation}.
First, we present the StyleRes results which is our baseline model.
Our set-up includes a flow prediction and warping the residual features. 
One straightforward way to obtain these flow predictions is to go to the image space for edited and unedited $W^+$ and calculate the flow with a pretrained flow estimation network on the image space. 
We first experiment with this set-up.
The flow predictions are estimated by \cite{truong2021warp} and downsampled to use in warping on the estimated residual features.
Since this set-up has to generate images with StyleGAN, it is computationally more expensive. 
We observe that one significant improvement is obtained in identity preservation as measured in Id scores.
That is because during the edit, with flow estimation, facial details are able to be carried to the output image. 

Next, we merge the flow prediction to our pipeline in an efficient way. Truong et. al. \cite{truong2021warp} estimates flow by first extracting VGG features \cite{simonyan2014very} from images. 
Instead, we use StyleGAN features that are generated mid-level for both edited and unedited features.
The flow estimation is again used for warping the residual features. 
The flow estimation can be guided via the image reconstruction and adversarial losses we use. 
However, it is difficult to learn the flow in this way, so we also experiment by adding the flow estimation loss.
It achieves even better results in the set-up where pretrained flow estimation is operated on image space while also achieving better run-time. 
That is because the flow prediction is now merged into the network and in end-to-end training, it can detect better features by specializing in this task.
We also provide results for the setting which does not use flow loss, the guidance in the training with pseudo-ground truths estimated by the pretrained flow estimation network.
Since it is difficult to train the flows via image reconstruction and adversarial losses, we observe a large benefit of the flow loss in our training.

Lastly, we experiment with different baseline models and compare StyleRes and WarpRes.
Previous experiments in the ablation study and final model use e4e following StyleRes and HFGI.
In this section, we experiment with other popular encoders that predict $W^+$ vectors from images namely, pSp and StyleTransformer. The quantitative and qualitative results are provided in Table \ref{table:ablation} and Fig. \ref{fig:results_bases}, respectively.
WarpRes significantly improves all encoders and can work with any base encoders we tested as shown in Table \ref{table:ablation}.
Its improvements over StyleRes are also significant and consistent.

We provide the base encoders editing results and when they are combined with WarpRes in Fig. \ref{fig:results_bases}.
Encoders miss many image details and reconstruct the images with low fidelity
while WarpRes is able to provide the details and correct the image details. For example, in the first two rows, encoders alone generate images with
different eye-glasses while WarpRes is able to correct them and additionally add wrinkles correctly which helps to preserve the identity and boost the identity score.
In the third example, encoders are not able to reconstruct the hair accessorize and pSp and StyleTransformer confuse them with hair and generate artifacts. WarpRes corrects the hair and reconstructs the hair accessorize. Similarly, in example 5, the microphone causes confusion in the encoders, especially for pSp and StyleTransformer, and results in dark regions in the chin which again is corrected when WarpRes is used. 
Note that different encoders encode images into different $W^+$ spaces and therefore result in slightly different edits. 
For example, StyleTransformer under-age edits also add eyeglasses and WarpRes relies on a base encoder for the editing spaces. Adding eyeglasses when age is edited is very common because of the correlations between these features \cite{shen2020interpreting}.

\textbf{Results.} We provide the final comparisons on CelebA-HQ and Stanford car datasets in Table \ref{table:results_face} and \ref{table:results_car}, respectively.
We achieved better results than the previous state-of-the-art by a large margin in all different metrics. 
In the reconstruction of original images, very large improvements are obtained in Table \ref{table:results_face}. 
Additionally, the identity preservation improvements are quite significant. For example, for pose change, the Id score is improved to $0.74$ compared to $0.68$ which is the best previous score obtained by HyperStyle. 
That is because, during the edits, WarpRes is able to move the high-frequency details correctly to the edited feature space.
Similarly, in cars, WarpRes obtains better scores by large margins both for reconstruction without edits and quality of images under edits. 
We also provide the inference run-time in Table \ref{table:results_face}.
With the addition of flow estimation, the method becomes slower than StyleRes, however, it still runs fast enough to make edits in real time. It is significantly faster than Restyle,  HyperStyle, and FeatureStyle.

% comparison images
% smile removal
\begin{figure*}
\centering
\scalebox{0.71}{
\addtolength{\tabcolsep}{-5pt}   
\begin{tabular}{cccccccccc}
\rotatebox{90}{~~~~~~~1.  Pose (-)} &
\interpfigt{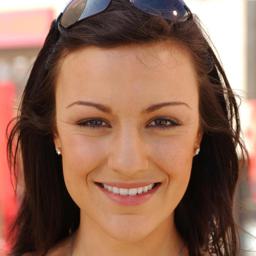} &
\interpfigt{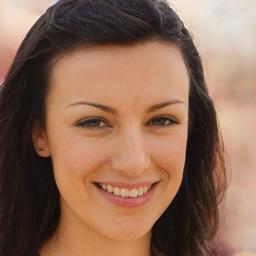} &
\interpfigt{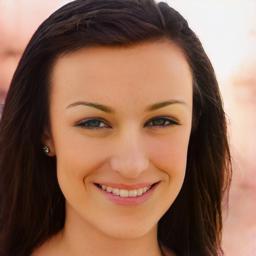} &
\interpfigt{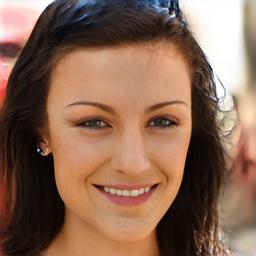} &
\interpfigt{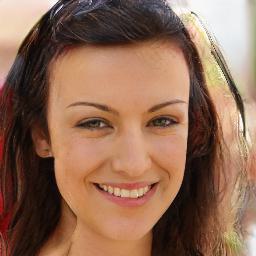} &
\interpfigt{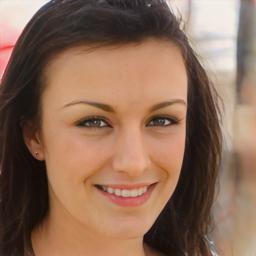} &
\interpfigt{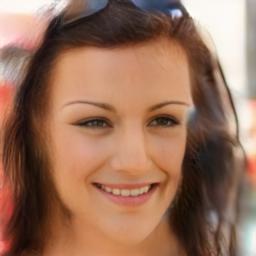} &
\interpfigt{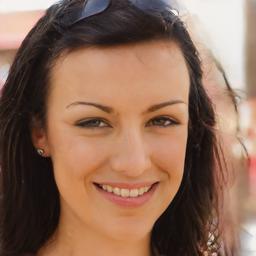} &
\interpfigt{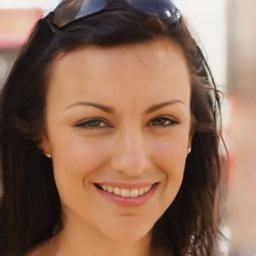}
\\

\rotatebox{90}{~~~~~~~2.  Pose (-)} &
\interpfigt{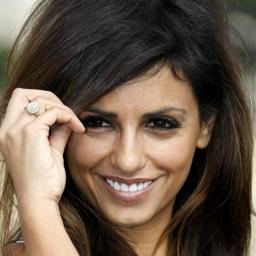} &
\interpfigt{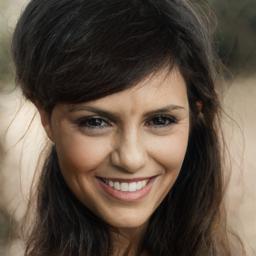} &
\interpfigt{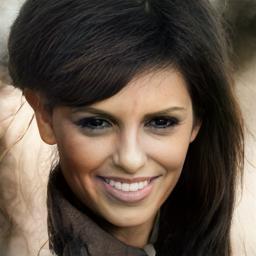} &
\interpfigt{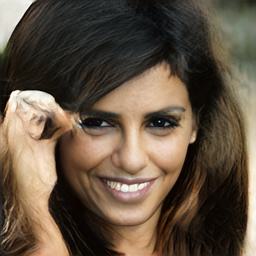} &
\interpfigt{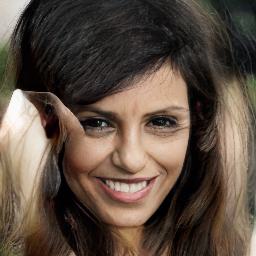} &
\interpfigt{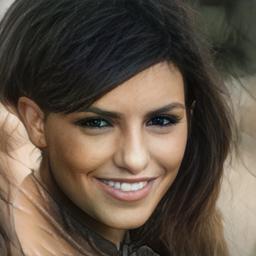} &
\interpfigt{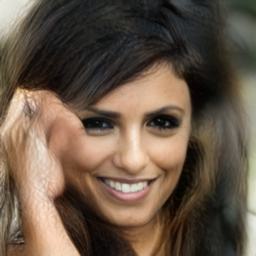} &
\interpfigt{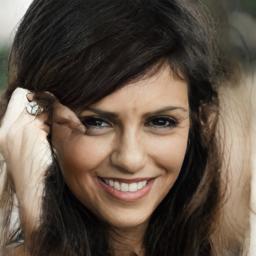} &
\interpfigt{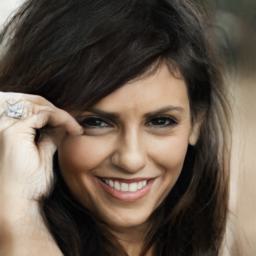}
\\
\rotatebox{90}{~~~~~~~3.  Pose (-)} &
\interpfigt{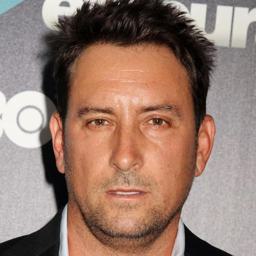} &
\interpfigt{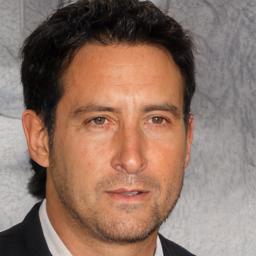} &
\interpfigt{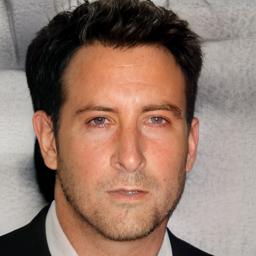} &
\interpfigt{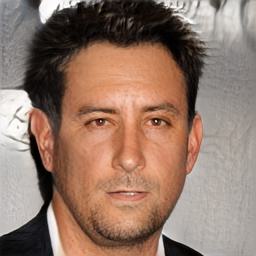} &
\interpfigt{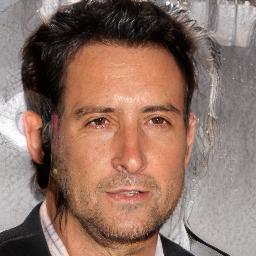} &
\interpfigt{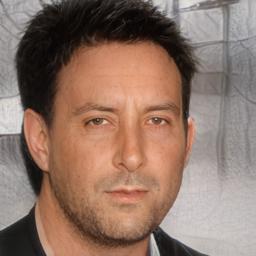} &
\interpfigt{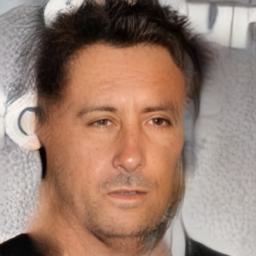} &
\interpfigt{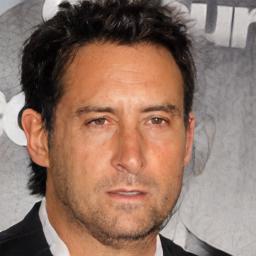} &
\interpfigt{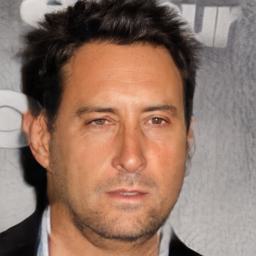}
\\

\rotatebox{90}{~~~~~~~4.  Pose (-)} &
\interpfigt{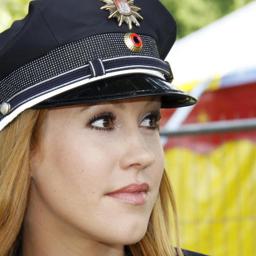} &
\interpfigt{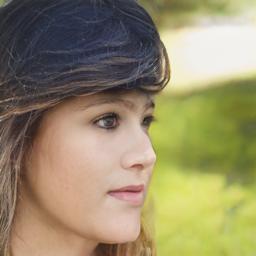} &
\interpfigt{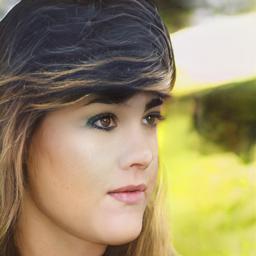} &
\interpfigt{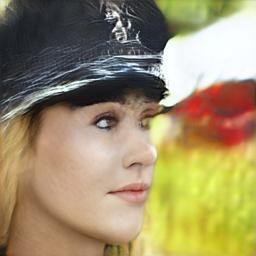} &
\interpfigt{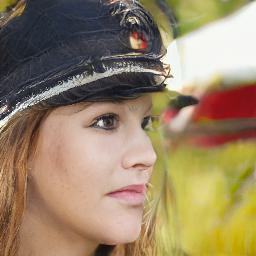} &
\interpfigt{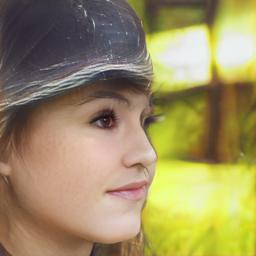} &
\interpfigt{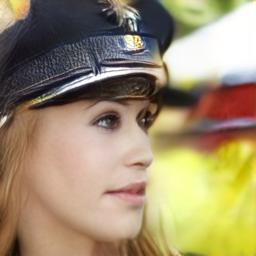} &
\interpfigt{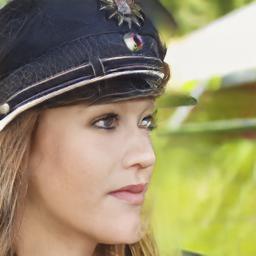} &
\interpfigt{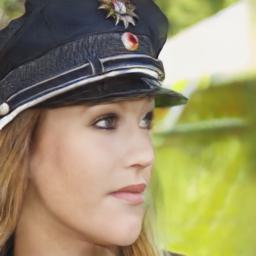}
\\
\rotatebox{90}{~~~~~~~5.  Pose (-)} &
\interpfigt{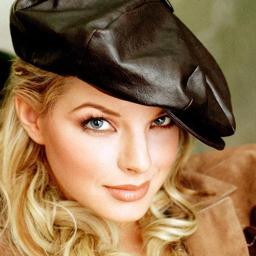} &
\interpfigt{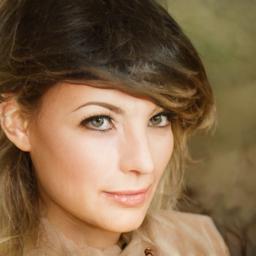} &
\interpfigt{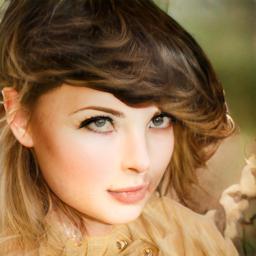} &
\interpfigt{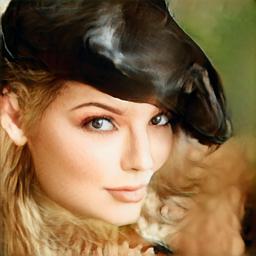} &
\interpfigt{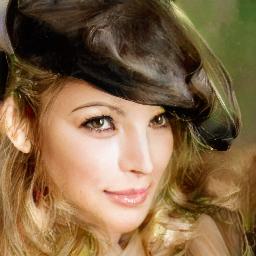} &
\interpfigt{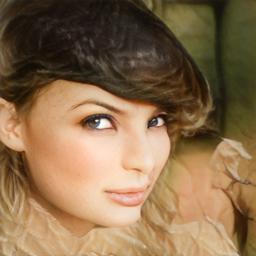} &
\interpfigt{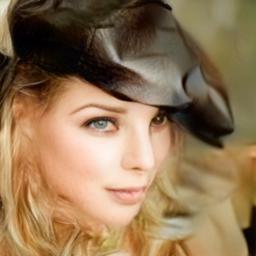} &
\interpfigt{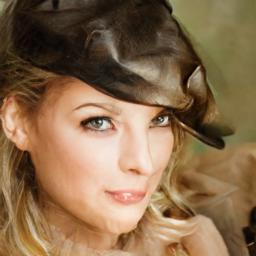} &
\interpfigt{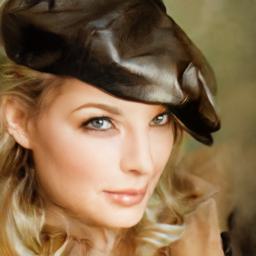}
\\

\rotatebox{90}{~~~~~~~6.  Pose (-)} &
\interpfigt{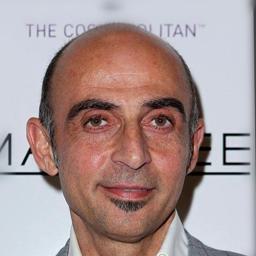} &
\interpfigt{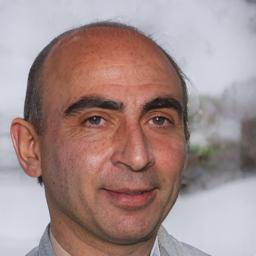} &
\interpfigt{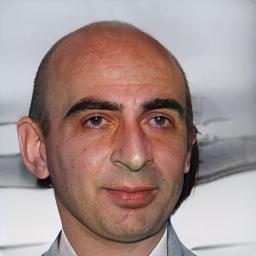} &
\interpfigt{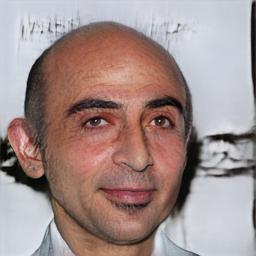} &
\interpfigt{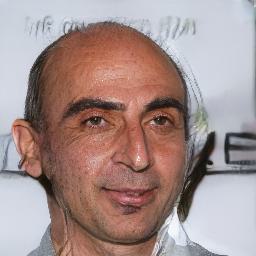} &
\interpfigt{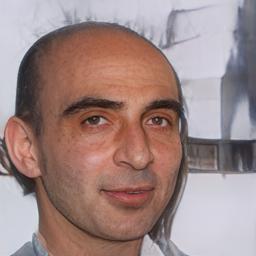} &
\interpfigt{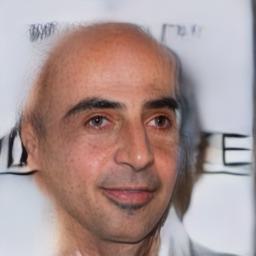} &
\interpfigt{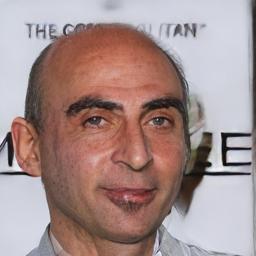} &
\interpfigt{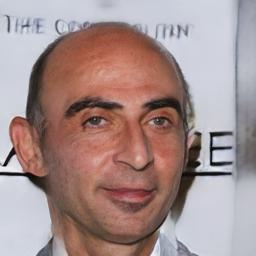}
\\

\rotatebox{90}{~~~~~~~7.  Smile (+)} &
\interpfigt{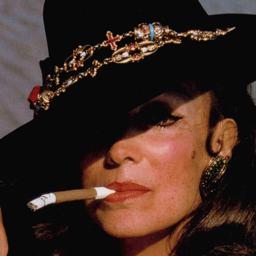} &
\interpfigt{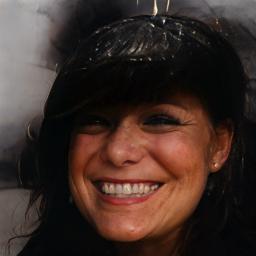} &
\interpfigt{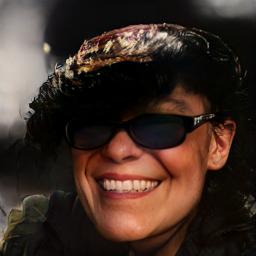} &
\interpfigt{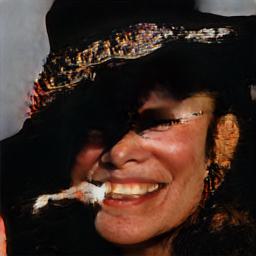} &
\interpfigt{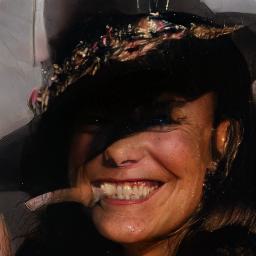} &
\interpfigt{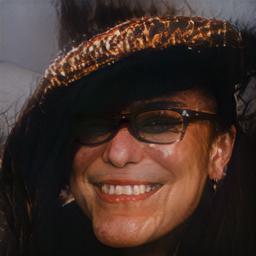} &
\interpfigt{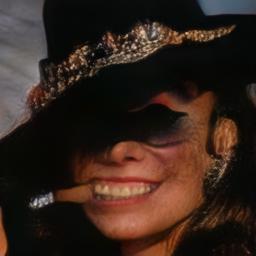} &
\interpfigt{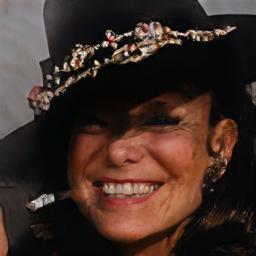} &
\interpfigt{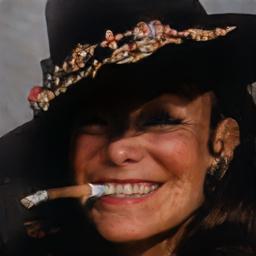}
\\
\rotatebox{90}{~~~~~~~8.  Smile (+)} &
\interpfigt{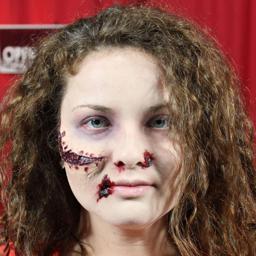} &
\interpfigt{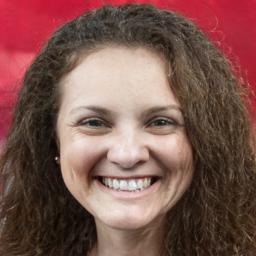} &
\interpfigt{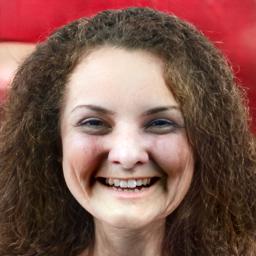} &
\interpfigt{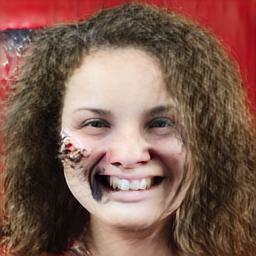} &
\interpfigt{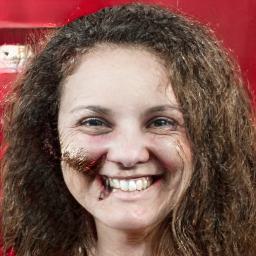} &
\interpfigt{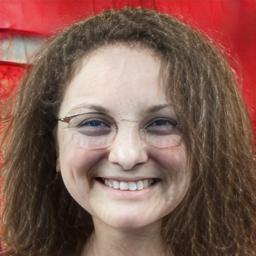} &
\interpfigt{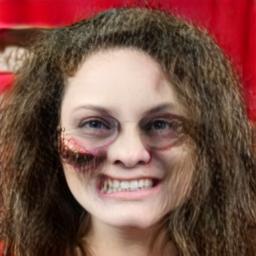} &
\interpfigt{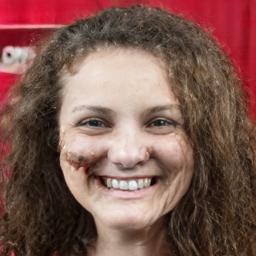} &
\interpfigt{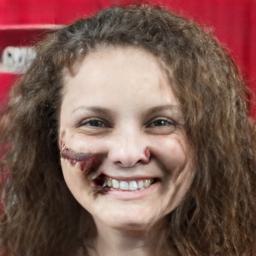}
\\

\rotatebox{90}{~~~~~~~9.  Smile (+)} &
\interpfigt{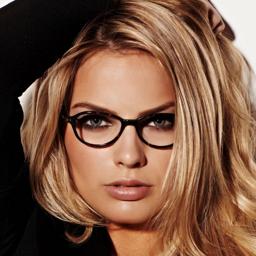} &
\interpfigt{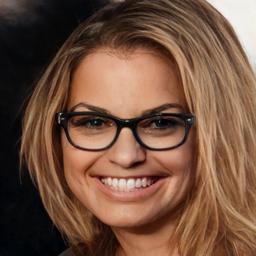} &
\interpfigt{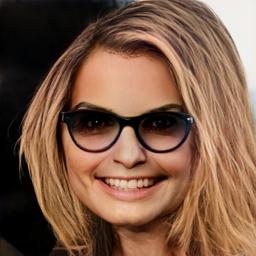} &
\interpfigt{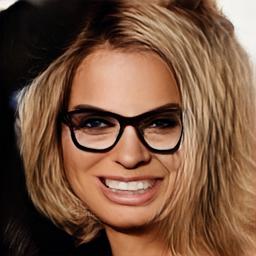} &
\interpfigt{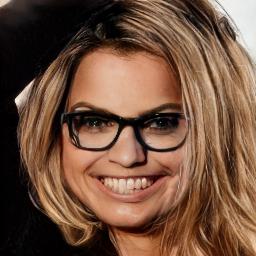} &
\interpfigt{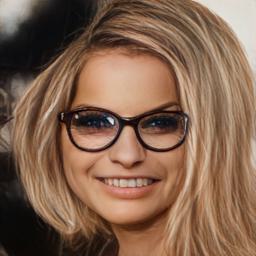} &
\interpfigt{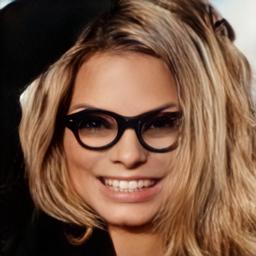} &
\interpfigt{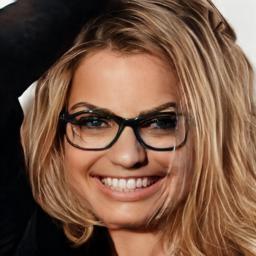} &
\interpfigt{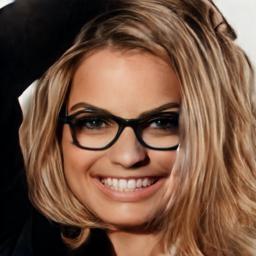}
\\
\rotatebox{90}{~~~~~~~10.  Age (-)} &
\interpfigt{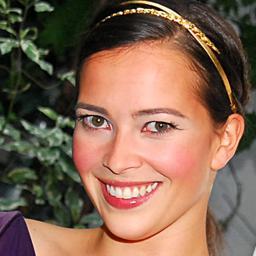} &
\interpfigt{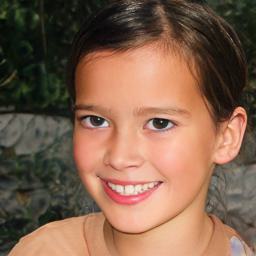} &
\interpfigt{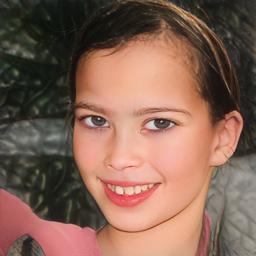} &
\interpfigt{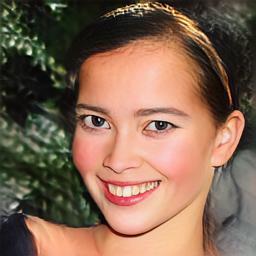} &
\interpfigt{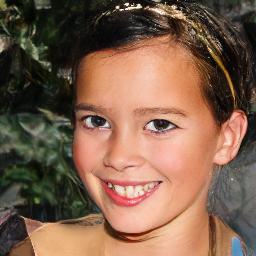} &
\interpfigt{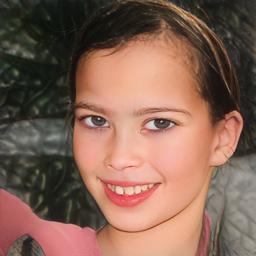} &
\interpfigt{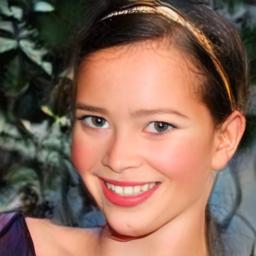} &
\interpfigt{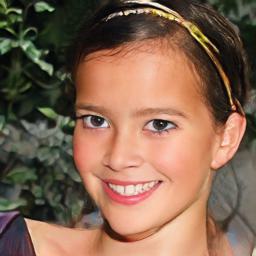} &
\interpfigt{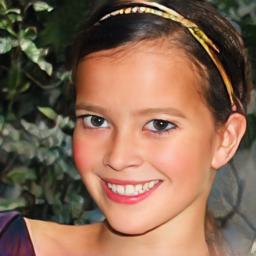}
\\
\rotatebox{90}{~~~~~~~11.  Age (-)} &
\interpfigt{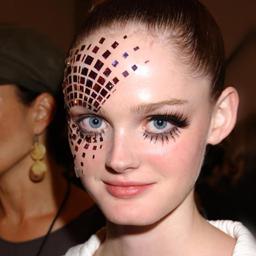} &
\interpfigt{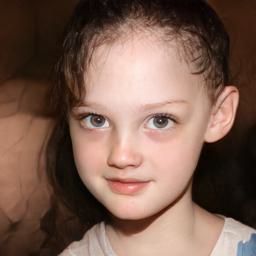} &
\interpfigt{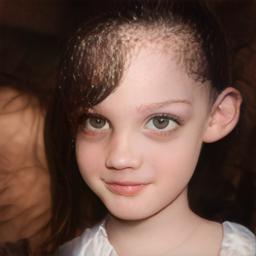} &
\interpfigt{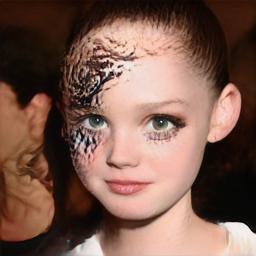} &
\interpfigt{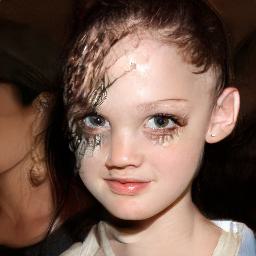} &
\interpfigt{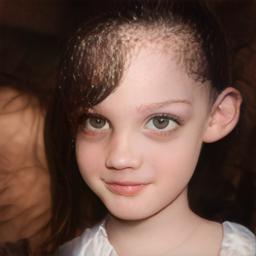} &
\interpfigt{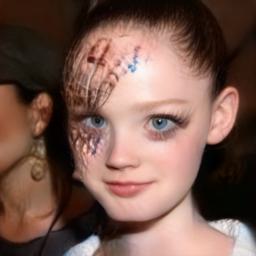} &
\interpfigt{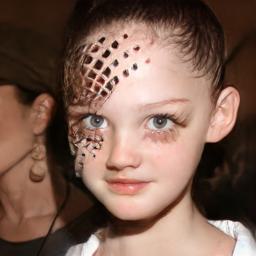} &
\interpfigt{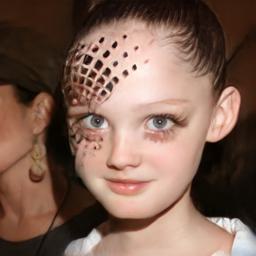}
\\

& Input &  e4e & ReStyle & HyperStyle & HFGI  & StyleTransformer & FeatureStyle & StyleRes & WarpRes \\
\end{tabular}
}
\caption{Qualitative results of our and competing methods on the CelebA-HQ dataset. Please refer to the text for discussion.}
\label{fig:results_ffhq}
\end{figure*}

In Fig. \ref{fig:results_ffhq}, we provide qualitative results on the CelebA-HQ dataset. We mostly share the most challenging edits which are struggled the most by the previous methods such as pose and smile. 
Previously, StyleRes provided large improvements over previous methods, however still struggles in the challenging scenarios, especially in pose edits.
For example, in the first and third rows, there are artifacts at the pixels that should belong to the background after the pose is changed. The second row shows an example where the hand is not reconstructed and edited correctly and causes artifacts on the edited images while WarpRes is the only one outputting a realistic edit image with high fidelity to the input.
In the fourth and fifth examples, WarpRes is the only one that correctly reconstructs the hats, hair, and eyes. 
The smile edit is the other challenging one because it also requires features to move spatially.
Similarly, in the eighth example, the facial makeup is not correctly warped when the image is edited for smile in the previous methods. 
Also, other artifacts exist in the seventh and ninth examples as well as in the outputs of previous methods, especially around cigarettes and eyeglasses. 
The age edit is easier since it usually does not require to warp features. Mostly, this edit requires making the chin more oval and that may cause artifacts as in the tenth example for HFGI and StyleRes while the other previous methods do not reconstruct the image faithfully. WarpRes is the only one that achieves high-fidelity reconstruction and high-quality editing. The last example has a face painting which causes a different challenge. Again WarpRes achieves significantly better results than the others.

\newcommand{\interpfigc}[1]{\includegraphics[trim=0 0 0cm 0, clip, width=3.0cm]{#1}}
\begin{figure*}
\centering
\scalebox{0.71}{
\addtolength{\tabcolsep}{-5pt}   
\begin{tabular}{ccccccccc}
\rotatebox{90}{~~1. Viewpoint} &
\interpfigc{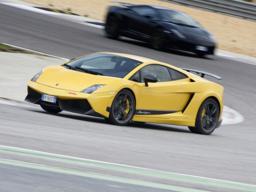} &
\interpfigc{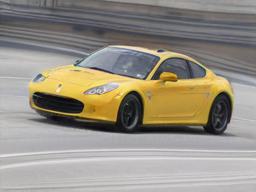} &
\interpfigc{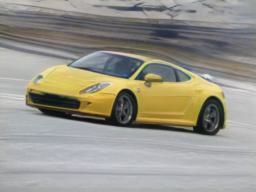} &
\interpfigc{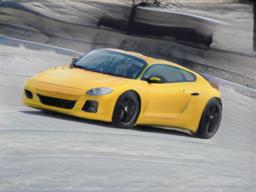} &
\interpfigc{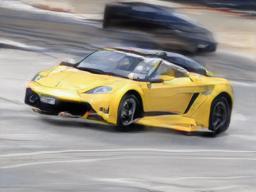} &
\interpfigc{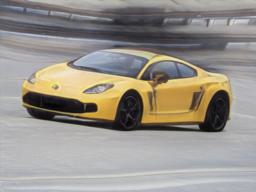} &
\interpfigc{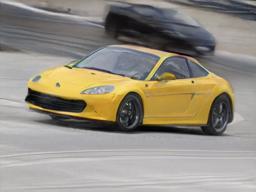} &
\interpfigc{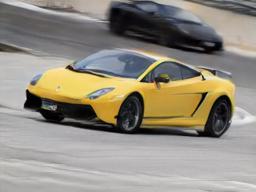}
\\
\rotatebox{90}{~~2. Viewpoint} &
\interpfigc{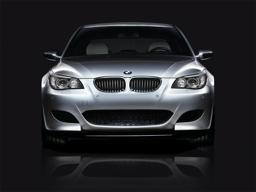} &
\interpfigc{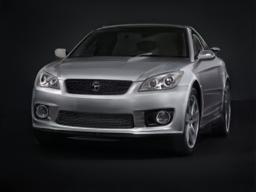} &
\interpfigc{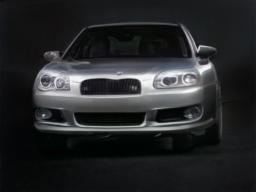} &
\interpfigc{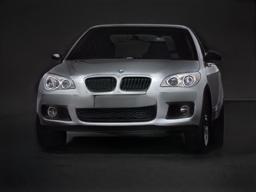} &
\interpfigc{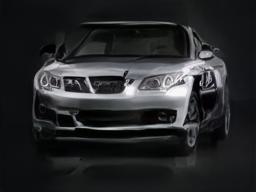} &
\interpfigc{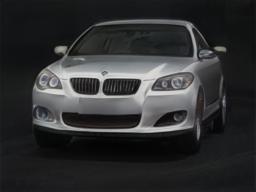} &
\interpfigc{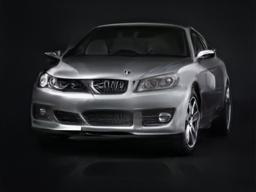} &
\interpfigc{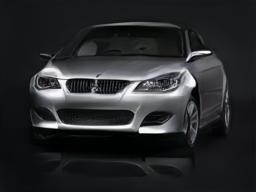}
\\
\rotatebox{90}{~~3. Viewpoint} &
\interpfigc{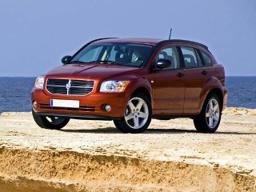} &
\interpfigc{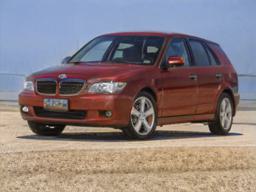} &
\interpfigc{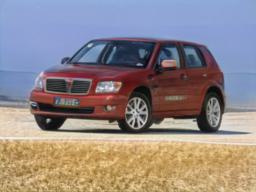} &
\interpfigc{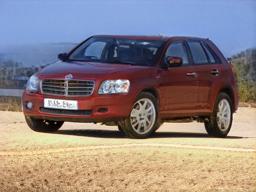} &
\interpfigc{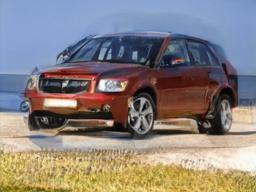} &
\interpfigc{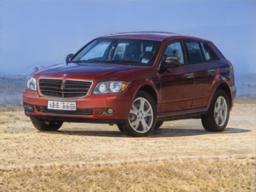} &
\interpfigc{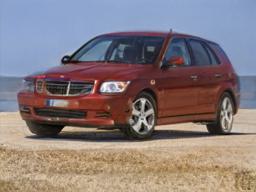} &
\interpfigc{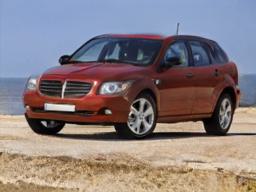}
\\
\rotatebox{90}{~~4. Viewpoint} &
\interpfigc{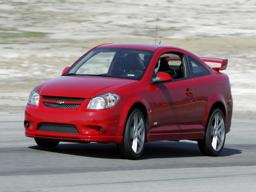} &
\interpfigc{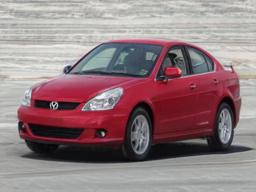} &
\interpfigc{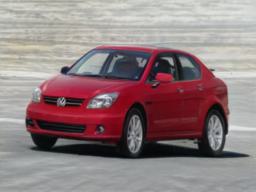} &
\interpfigc{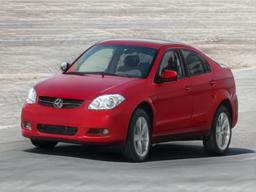} &
\interpfigc{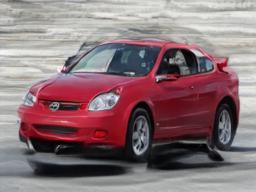} &
\interpfigc{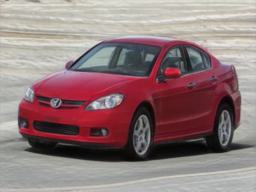} &
\interpfigc{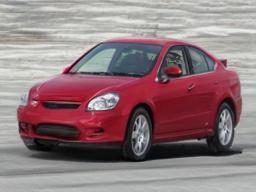} &
\interpfigc{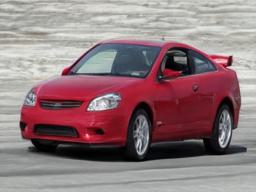}
\\
\rotatebox{90}{~~~~~~~5.  Color} &
\interpfigc{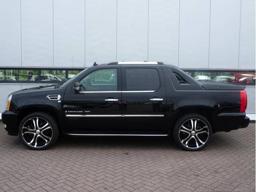} &
\interpfigc{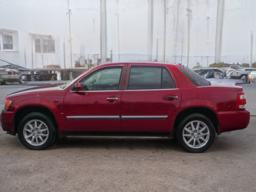} &
\interpfigc{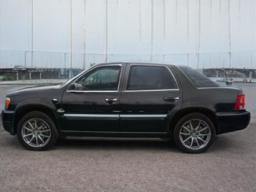} &
\interpfigc{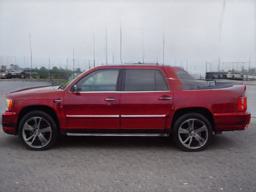} &
\interpfigc{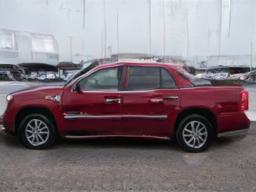} &
\interpfigc{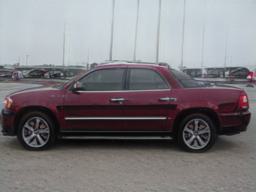} &
\interpfigc{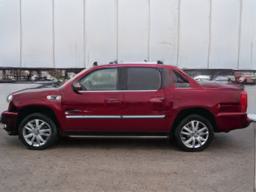} &
\interpfigc{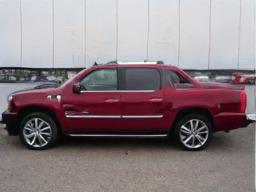}
\\
\rotatebox{90}{~~~~~~~6.  Color} &
\interpfigc{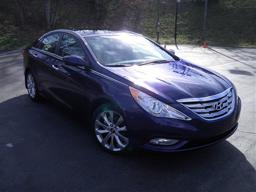} &
\interpfigc{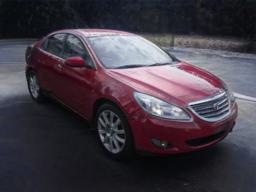} &
\interpfigc{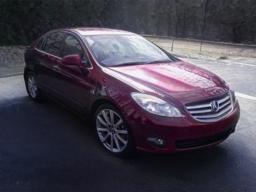} &
\interpfigc{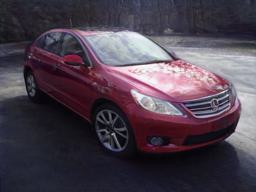} &
\interpfigc{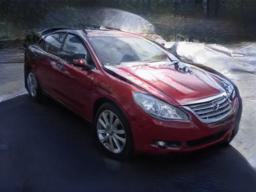} &
\interpfigc{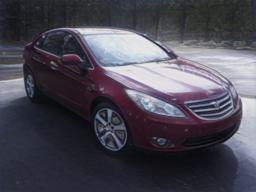} &
\interpfigc{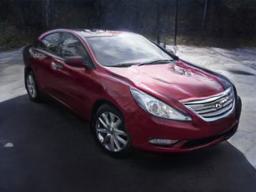} &
\interpfigc{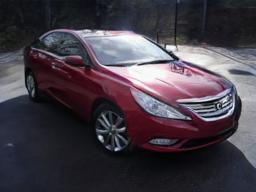}
\\
\rotatebox{90}{~~~~~~~7. Color} &
\interpfigc{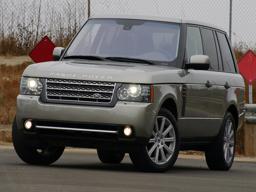} &
\interpfigc{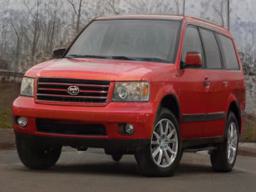} &
\interpfigc{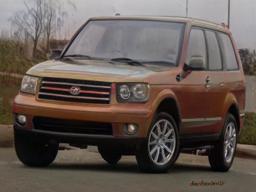} &
\interpfigc{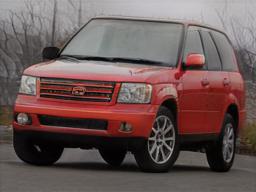} &
\interpfigc{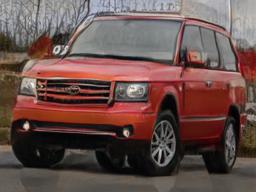} &
\interpfigc{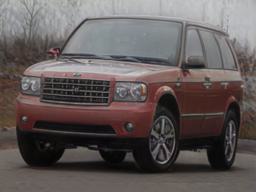} &
\interpfigc{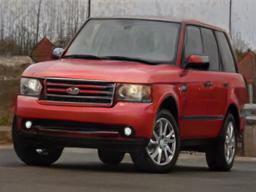} &
\interpfigc{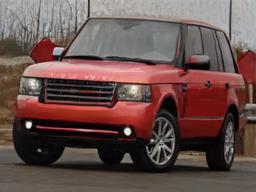}
\\
\rotatebox{90}{~~~~~~~8. Color} &
\interpfigc{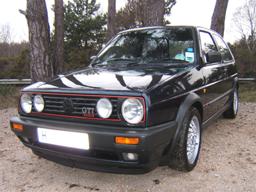} &
\interpfigc{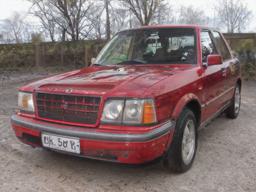} &
\interpfigc{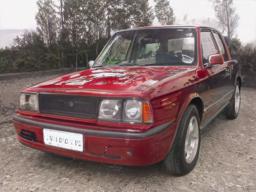} &
\interpfigc{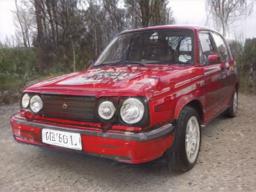} &
\interpfigc{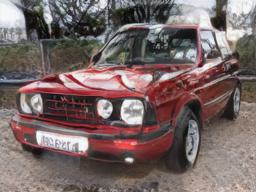} &
\interpfigc{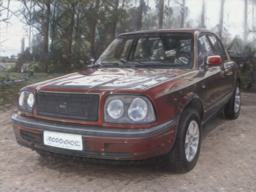} &
\interpfigc{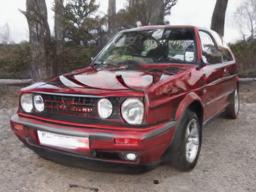} &
\interpfigc{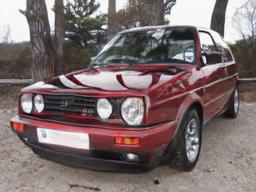}
\\
\rotatebox{90}{~~~~~~~9. Grass} &
\interpfigc{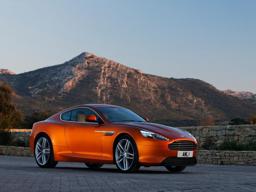} &
\interpfigc{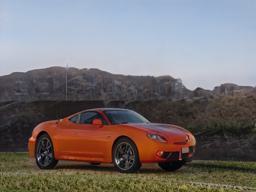} &
\interpfigc{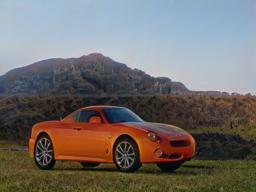} &
\interpfigc{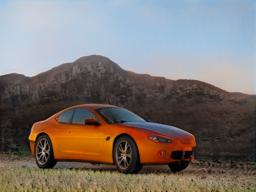} &
\interpfigc{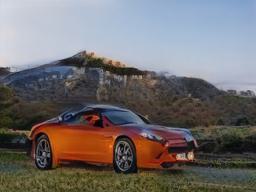} &
\interpfigc{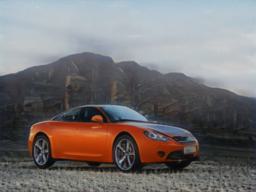} &
\interpfigc{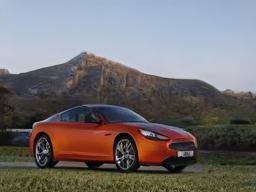} &
\interpfigc{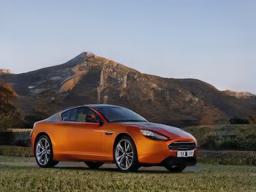}
\\
\rotatebox{90}{~~~~~10.  Grass} &
\interpfigc{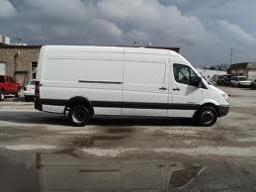} &
\interpfigc{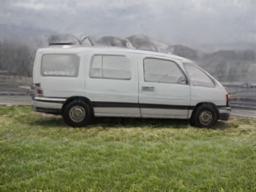} &
\interpfigc{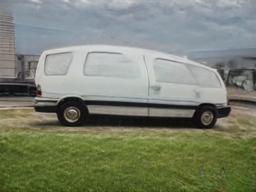} &
\interpfigc{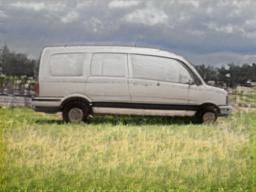} &
\interpfigc{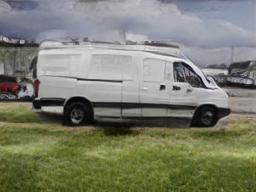} &
\interpfigc{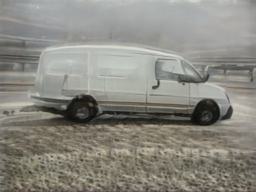} &
\interpfigc{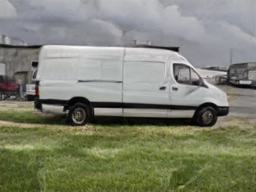} &
\interpfigc{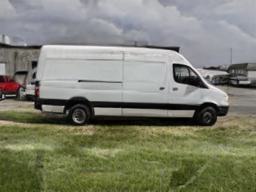}
\\
\rotatebox{90}{~~~~11.  Grass} &
\interpfigc{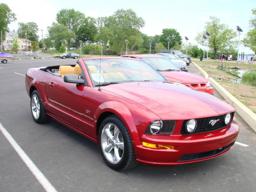} &
\interpfigc{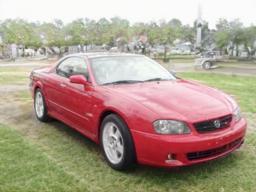} &
\interpfigc{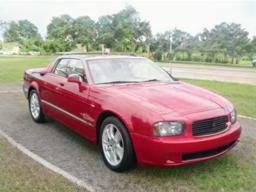} &
\interpfigc{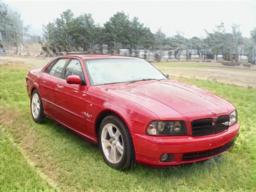} &
\interpfigc{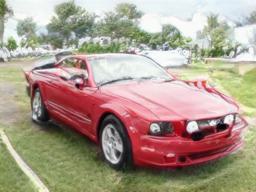} &
\interpfigc{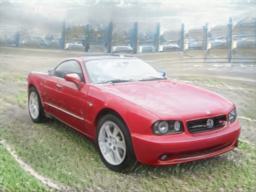} &
\interpfigc{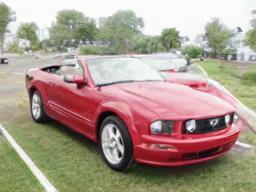} &
\interpfigc{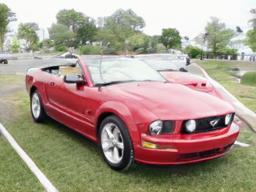}
\\
& Input &  e4e & ReStyle & HyperStyle & HFGI  & StyleTransformer & StyleRes & WarpRes \\
\end{tabular}
}
\caption{Qualitative results of our and competing methods on the Stanford cars dataset. Please refer to the text for discussion.}
\label{fig:results_car}
\end{figure*}

In Fig. \ref{fig:results_car}, we provide qualitative results on the Stanford cars dataset. 
First, we show the edit results of changing viewpoints such as rotating and zooming. 
This is a challenging edit and methods that reconstruct high-rate latent codes such as HFGI and StyleRes and ours need to transform the features based on the viewpoint change.
In the first four examples, e4e, ReStyle, HyperStyle, and StyleTransformer fail to keep the identity of the object the same. On the other hand, HFGI results in severe artifacts because of the high rate of latent codes not adopting correctly to the edits. StyleRes results in better results but it also removes many image details instead of correctly transforming them. 
WarpRes achieves warping the high-rate details of the cars to the correct locations in the feature space. 

\newcommand{\interpfigd}[1]{\includegraphics[trim=0 0 0cm 0, clip, width=2.3cm]{#1}}
\begin{figure*}
\centering
\scalebox{0.71}{
\addtolength{\tabcolsep}{-5pt}   
\begin{tabular}{ccccccccccc}
\interpfigd{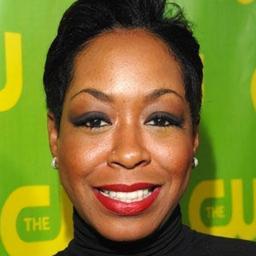} &
\rotatebox{90}{~~~WarpRes} &
\interpfigd{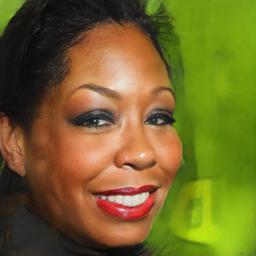} &
\interpfigd{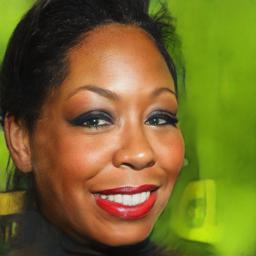} &
\interpfigd{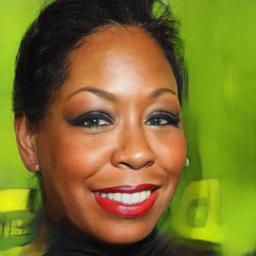} &
\interpfigd{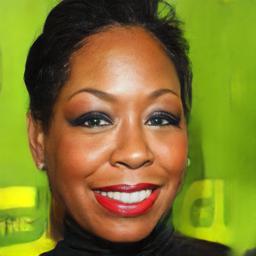} &
\interpfigd{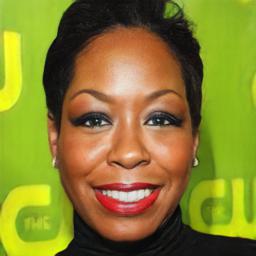} &
\interpfigd{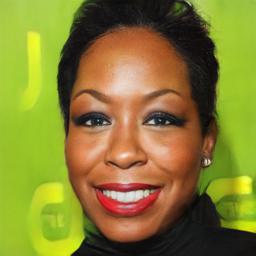} &
\interpfigd{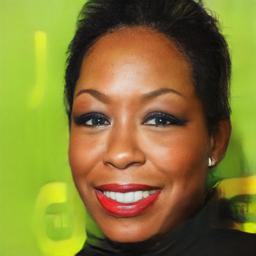} &
\interpfigd{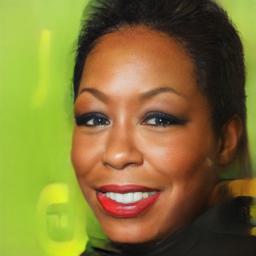} &
\interpfigd{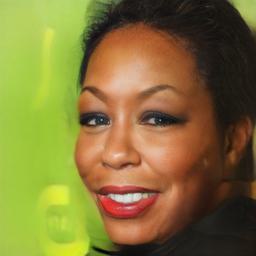} \\
& \rotatebox{90}{~~~Eg3d-goae} &
\interpfigd{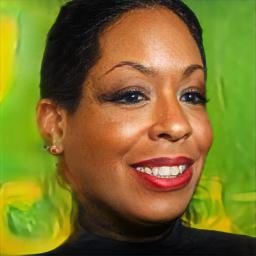} &
\interpfigd{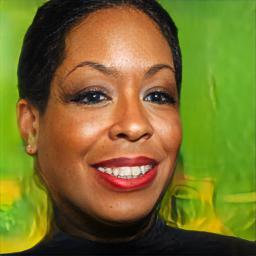} &
\interpfigd{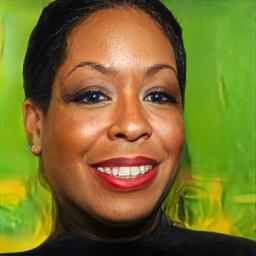} &
\interpfigd{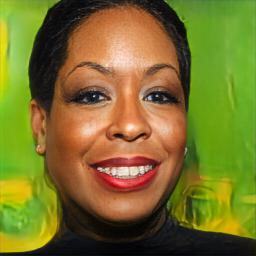} &
\interpfigd{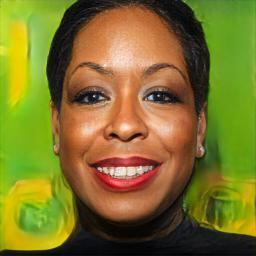} &
\interpfigd{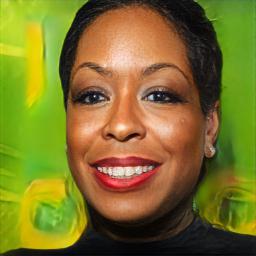} &
\interpfigd{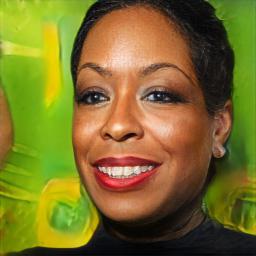} &
\interpfigd{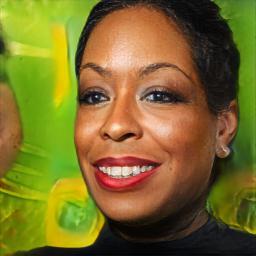} &
\interpfigd{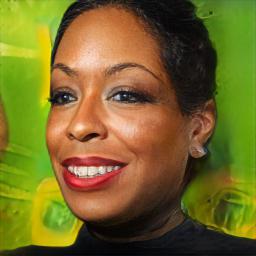} \\
\interpfigd{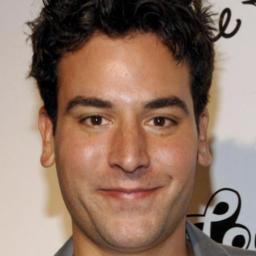} &
 \rotatebox{90}{~~~WarpRes} &
\interpfigd{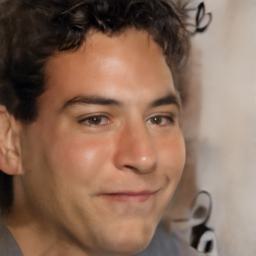} &
\interpfigd{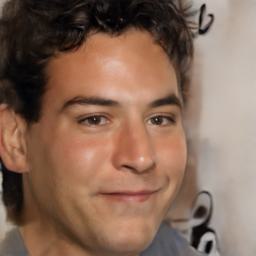} &
\interpfigd{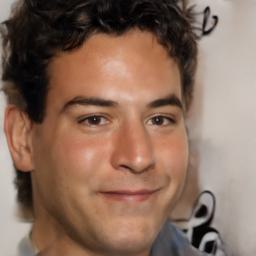} &
\interpfigd{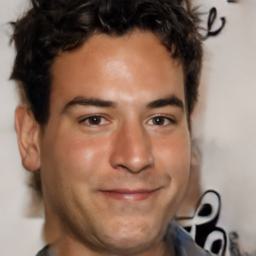} &
\interpfigd{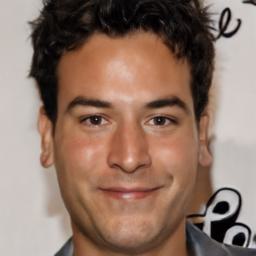} &
\interpfigd{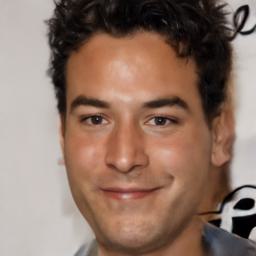} &
\interpfigd{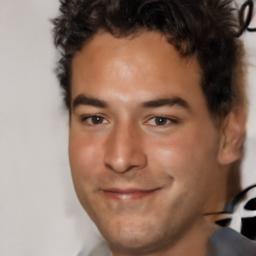} &
\interpfigd{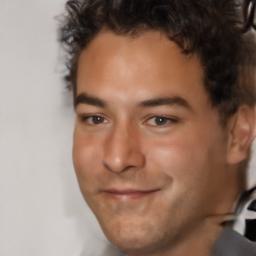} &
\interpfigd{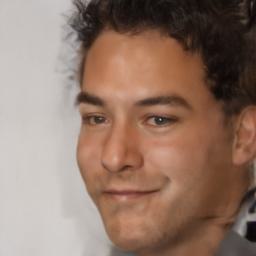} \\
& \rotatebox{90}{~~~Eg3d-goae} &
\interpfigd{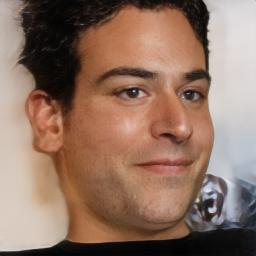} &
\interpfigd{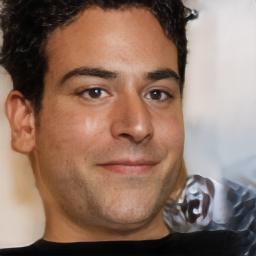} &
\interpfigd{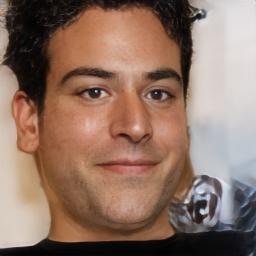} &
\interpfigd{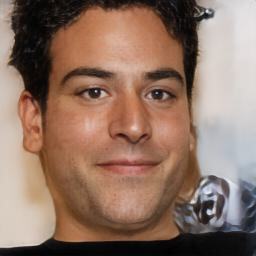} &
\interpfigd{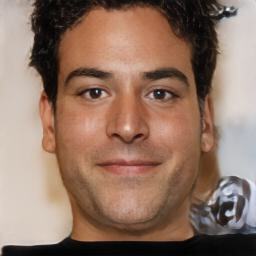} &
\interpfigd{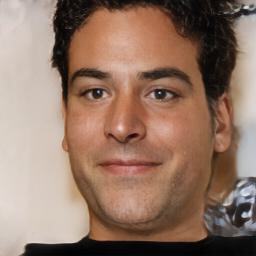} &
\interpfigd{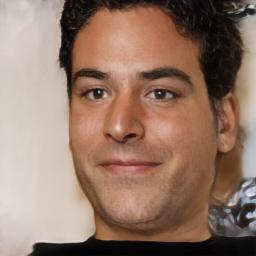} &
\interpfigd{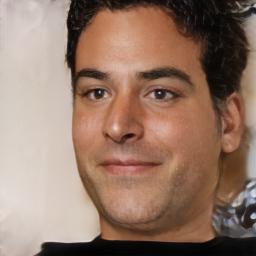} &
\interpfigd{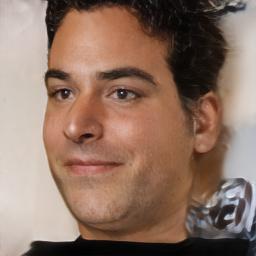} \\
\end{tabular}
}
\caption{Comparison of 2d (StyleGAN) and 3d (Eg3d) GAN inversion methods, WarpRes and Eg3d-goae, respectively.}
\label{fig:results_3d}
\end{figure*}

Edits like color change and grass addition are simpler than viewpoint edits because they do not require transforming or warping the features.
However, previous inversion methods still struggle with these edits. StyleRes achieves better results but when zoomed in, artifacts exist in its outputs as well. 
For example, in the fifth example, StyleRes cannot generate the door handles while WarpRes can. 
In the sixth and seventh examples, WarpRes is the only method that generates the tires and the front bumper with high fidelity to the input, respectively. 
Similarly, in the grass addition edits, other methods struggle and miss many realistic details compared to WarpRes.

\textbf{Comparisons with 3D-aware GAN Inversion Models.} Recently, 3D-aware GAN models have been proposed \cite{gao2022get3d, chan2022efficient} to achieve view-point manipulations of generations while preserving the identity.  3D-aware GANs include a 3D-structure-aware inductive bias in the generator network architecture and a neural renderer to achieve training of the network parameters.
Image inversion methods \cite{e3dge, hfgi3d, ide_3d, yuan2023make} are also proposed for 3D-aware GAN models especially for the efficient Eg3d model \cite{chan2022efficient}.
In Fig. \ref{fig:results_3d}, we compare the state-of-the-art inversion method of the Eg3d model which is Eg3d-goae \cite{yuan2023make} and WarpRes to discuss their advantages and disadvantages.
Firstly, 3D-aware GAN models provide a promising direction to achieve view-point edits. 
However, their image generation realism is not on par with StyleGAN models yet  \cite{gao2022get3d, chan2022efficient}. 
Secondly, the inversion methods are not able to map images to a latent space that will result in high-detailed reconstruction of the input. 
It may be because of the limitations of the generator or the inversion methods since it is a newer research area.
WarpRes combined with StyleGAN achieves better fidelity to the input image and is able to change the viewpoint as shown in Fig. \ref{fig:results_3d}. 
It is especially significantly better when the rotations are smaller as in the three examples from the center.
On the other hand, WarpRes with StyleGAN cannot promise strict multi-view consistency since it does not output an explicit 3D model which is a limitation of this work.
Note that we do not compare the methods quantitatively because the 2D inversion pose edit is relative to the original pose whereas, with Eg3d, the edit is applied by changing the camera parameters. It is not possible to align these two methods of editing automatically.
We manually pick the edit parameters to align the rotations in Fig. \ref{fig:results_3d} for these two examples.

\newcommand{\interpfigf}[1]{\includegraphics[trim=0 0 0cm 0, clip, width=2.6cm]{#1}}
\begin{figure}
\centering
\scalebox{0.71}{
\addtolength{\tabcolsep}{-5pt}   
\begin{tabular}{cccc}
\interpfigf{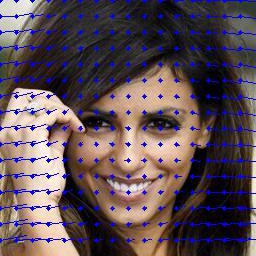} &
\interpfigf{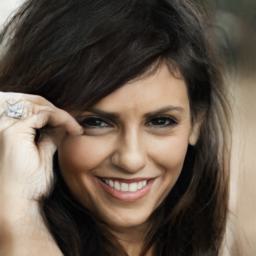} & 
\interpfigf{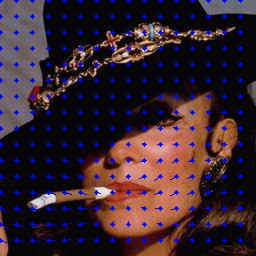} &
\interpfigf{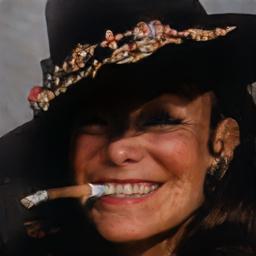} \\
Pose flow & Output & Smile Flow & Output \\
\interpfigf{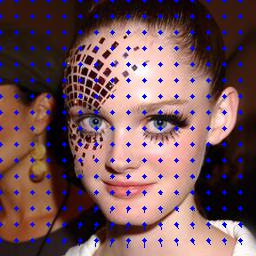} &
\interpfigf{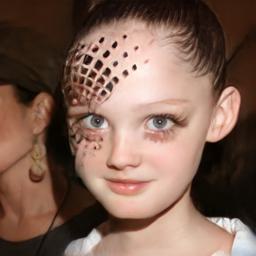} & 
\interpfigf{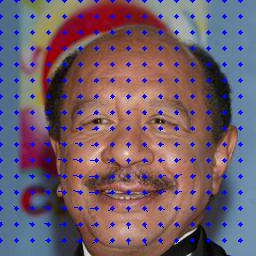} &
\interpfigf{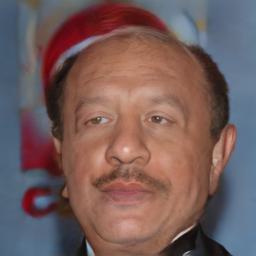} \\
Young flow & Output & Smile Flow & Output \\
\end{tabular}
}
\caption{Visualization of flow estimation for different edits. }
\label{fig:results_flow}
\end{figure}

\textbf{Visualization of flow estimations.} In Fig. \ref{fig:results_flow}, we visualize the flow estimations for different edits. We visualize the flows estimated at $64\times64$ dimension. We upsample and multiply the flow values by $4$ to visualize them on the input image. 
Flows are correctly predicted to enable warping the features to their correct place based on the edits. 
For example, when the smile is added, the flows are outward from the mouth whereas when the smile is removed, they are inward.
\section{Conclusion}

This paper proposes a GAN inversion method that is run-time efficient and achieves high-fidelity and high-quality image editing under editing directions that are explored in various works.
Different than previous works, we propose to estimate flow predictions for unedited and edited features to warp high-rate residual features that are needed for precise image reconstruction.
We show that the proposed framework achieves state-of-the-art results on numerous edits, especially on the ones that require large transformations.

\ifCLASSOPTIONcaptionsoff
  \newpage
\fi

{
\bibliographystyle{ieee}
\bibliography{ref}
}

\end{document}